\documentclass[preprint,12pt,a4paper]{elsarticle}

\usepackage[a4paper,left=2cm,right=2cm,top=2.5cm,bottom=2.5cm]{geometry}
\usepackage{amsmath,amssymb}
\usepackage{graphicx}
\usepackage{booktabs}
\usepackage{tabularx}
\usepackage{adjustbox}
\usepackage{algorithm}
\usepackage{algpseudocode}
\usepackage[font=small,labelfont=bf]{caption}
\usepackage{hyperref}
\usepackage{placeins}

\begin{document}

\begin{frontmatter}

\title{NS-ST-GraphRAG: Neuro-Symbolic Spatio-Temporal GraphRAG for Literary Knowledge Processing}

\author[1]{Lin Zheng Kui}
\affiliation[1]{organization={Joint Laboratory for Port Big Data and Intelligent Applications, Dalian Ocean University, Dalian 116023, China}}
\ead{dalianjx@163.com}

\begin{abstract}
Long-form literary narratives pose a distinctive information-processing challenge for retrieval-augmented generation: relevant evidence is distributed across chapters, relations evolve over narrative time, and correct answers may depend jointly on temporal, spatial, and relational constraints. We propose NS-ST-GraphRAG, a neuro-symbolic spatio-temporal GraphRAG framework that integrates ontology-guided extraction, deterministic constraint checking, dual temporal coordinates, spatial scene attributes, and dynamic sub-graph retrieval. Instead of retrieving from a single corpus-level graph, the framework selects the graph state valid for the temporal and spatial scope of a query and grounds generated answers in traceable evidence. We further introduce Red-Chamber-QA, to our knowledge the first open multi-hop question-answering benchmark for classical Chinese literature. Its 104-question preliminary set and 120-question held-out split are constructed under deterministic shortcut controls and expert verification, with time-, space-, and general-question categories and per-part evidence spans. Evaluation follows a frozen, dual-track protocol combining strict mechanical answer reproduction, an independent semantic judge, and deterministic citation-faithfulness checks. On the held-out split, NS-ST-GraphRAG achieves mechanical answer reproduction of $0.733$, compared with $0.675$ for the frozen window baseline and $0.083$ for a closed-book model. The system-baseline difference is directionally favorable but not statistically significant (McNemar exact $p=0.092$), while semantic-judge accuracy is $0.866$ versus $0.850$, with no reliable difference. The pre-specified constrained-category condition of H2 is not supported by the delivered comparison. These results show how temporal graph representation, constrained extraction, and auditable evaluation can be integrated into a unified framework for verifiable knowledge processing over long-form narrative.
\end{abstract}


\end{frontmatter}

\section{Introduction}
Long-form literary narrative is a stress test for retrieval-augmented systems. A novel such as Dream of the Red Chamber runs to a hundred and twenty chapters and some four hundred named characters, and the questions a reader asks of it are inherently relational and situated: which servants served which household before a given event, how a character's standing among relatives changed across the story, who was present at a scene and why that mattered later. These questions are the natural interface to a literary corpus, and they are precisely the questions that modern retrieval pipelines are least equipped to handle. Chunk-based vector retrieval, the standard RAG baseline, is not explicitly designed to preserve narrative-time structure when supporting evidence is distributed across chapters \cite{lewis2020rag,gao2024survey}. Graph-augmented generation pipelines improve on this by indexing entities and relations \cite{edge2024graphrag,guo2024lightrag,gutierrez2024hipporag}, but their graph representations do not encode narrative-time validity intervals, so states before and after a narrative turning point are not explicitly separated. A query about the state of the network at a specific story event cannot be routed to the temporally correct subgraph, because the graph carries no validity intervals. The problem is not a niche concern of one novel; every long-form narrative, historical chronicle, and serialized corpus that unfolds over time poses the same question to retrieval systems, and standard graph-augmented pipelines generally do not encode narrative time explicitly.

Three gaps follow directly. First, static graph construction discards the temporal and spatial structure that long-form narrative explicitly carries, so existing GraphRAG systems cannot express topology evolution over the narrative timeline \cite{liang2024a,plamper2025stkg}. Second, unconstrained LLM extraction over metaphor-rich prose hallucinates relations that violate the story's internal logic, such as impossible kinship links or master-servant bonds across rival households; hallucination is a documented failure class \cite{huang2024surveyhalluc}, and the evaluation of such hallucinations lacks a shared protocol beyond atomic factual precision for generation \cite{min2023factscore}. The distinction matters because generation-level metrics do not see a relation that should never have entered the graph in the first place. Third, there is no open multi-hop QA benchmark for classical Chinese literature: narrative QA benchmarks exist for books in English \cite{kocisky2018narrativeqa}, and prior work on this corpus evaluates extraction on a small ad hoc gold set rather than a released question-answering benchmark \cite{yuan2026knowledge}. Without a benchmark, claims about graph-grounded answering on this literature cannot be compared across systems, and each new system reports numbers on its own private test set.

We propose NS-ST-GraphRAG, a framework with three coupled components. A neuro-symbolic constraint layer checks every extracted triple against a domain ontology of genealogical, master-servant, and title-consistency rules and repairs violations through a bounded feedback loop. A spatio-temporal graph model stores entities and relations with dual temporal coordinates, a coarse chapter index and a fine narrative-time event interval, plus spatial scene attributes, so the graph is versioned over the story. A dynamic sub-graph retrieval stage parses time and space constraints from a question, retrieves the matching narrative slice by index lookup, and grounds every answer in supporting triples whose coordinates can be traced. The three components are designed to be individually auditable: every rejected relation names its rule, every retrieved answer names its slice, every answer names its evidence. This auditability addresses a broader limitation in hallucination-mitigation research, where evaluation protocols and evidence of effectiveness remain heterogeneous across approaches \cite{wagner2025mitigating}.

The evaluation pairs the framework with Red-Chamber-QA, a multi-hop QA golden dataset we construct over Dream of the Red Chamber with dedicated time-constrained and space-constrained question categories, provenance annotations, and a human-expert difficulty baseline. The present paper reports the first measured installments (Section~4): the preliminary benchmark set and its baseline characterization under a dual-track protocol, the full-corpus construction installment with its development-set trajectory, and the first held-out campaign under the frozen configuration against the same frozen window baseline and a closed-book floor. The full comparison against Naive Vector RAG, GraphRAG, LightRAG, and HippoRAG under the prompt-equalization fairness protocol, with the extraction, hallucination, provenance, latency, and construction-cost metrics, remains committed in advance for the complete release. Because the benchmark and the protocol were committed before the full measurement, the reported results are comparable across systems by construction rather than by retrospective alignment.

The contributions of this paper are:
\begin{itemize}
\item a neuro-symbolic constraint layer that detects and corrects hallucinated entities and relations during LLM-based literary knowledge extraction, driven by an LLM-assisted, expert-verified domain ontology whose interpretable rules bound the extraction error space, in contrast to the extraction loops without an explicit ontology-based constraint-repair layer of \cite{edge2024graphrag,yuan2026knowledge};
\item a spatio-temporal knowledge graph model with a dual temporal coordinate system, coarse chapter index plus fine narrative-time event intervals, and spatial scene attributes, enabling topology-evolution tracking that graph representations without explicit narrative-time validity intervals cannot directly support \cite{guo2024lightrag,gutierrez2024hipporag};
\item the Red-Chamber-QA benchmark, to our knowledge the first multi-hop QA golden dataset for classical Chinese literature, surveyed across the digital humanities and NLP venues covered in the related work, with time- and space-constrained categories and provenance annotations: its 104-question preliminary set is publicly available under mechanical and expert verification together with a first measured baseline characterization, the implementation repository will be released open-source upon acceptance, and the full evaluation protocol covering four retrieval baselines, ablation, efficiency, and construction cost remains committed in advance for the complete release.
\end{itemize}

\section{Related Works}
\subsection{Retrieval-Augmented Generation over Structured Knowledge}
Retrieval-augmented generation grounds LLM output in retrieved evidence \cite{lewis2020rag}, and its survey literature establishes the standard design dimensions of retrieval, generation, and augmentation \cite{gao2024survey}. Our work sits on the structured-knowledge branch of this family. GraphRAG builds an entity-relation graph from the corpus, clusters the graph into communities, and summarizes each community so that global questions receive abstracted evidence \cite{edge2024graphrag}; LightRAG couples low-level entity-relation retrieval with high-level thematic retrieval over a graph index \cite{guo2024lightrag}; HippoRAG retrieves paths by personalized PageRank over OpenIE triples, drawing on the long lineage of open information extraction \cite{gutierrez2024hipporag,page1999pagerank,etzioni2011openie}; RAPTOR organizes documents into a recursive summarization tree whose upper nodes answer topical questions directly \cite{sarthi2024raptor}; and Self-RAG adds a self-reflection pass that critiques retrieved and generated content before answering \cite{asai2023selfrag}. The survey of graph-augmented generation records this convergence of graph structures and retrieval, and end-to-end relation extraction models supply the triples these systems consume \cite{peng2024graphragsurvey,huguetcabot2021rebel}. The graph-based systems above---GraphRAG, LightRAG, and HippoRAG---build their graphs without explicit narrative-time validity intervals, so a question about the state of a social network before and after a narrative turning point cannot be answered from the graph's own structure; the evidence for both sides is mixed in a single snapshot. This missing interval layer is the departure point of our spatio-temporal design.

\subsection{Knowledge Graphs for Literary Texts and Digital Humanities}
Constructing knowledge graphs from literary corpora has a lineage in computational literary studies. The foundational line extracts character social networks from dialogue interaction, inferring both speaker and addressee from quoted speech \cite{elson2010social}, and later work learns distributed story representations from those networks \cite{lee2020story}, under the distant-reading paradigm of quantitative literary analysis \cite{moretti2013distant}. Character identification in prose is itself a recognized hard problem because of aliases, honorifics, and indirect reference \cite{vala2015mrbennet}, which our entity-resolution stage confronts directly, and betweenness-style network measures supply the analytic vocabulary for the evolution questions the case study poses. On the classical Chinese corpus, prior work constructs a knowledge graph for the same novel we use, through a human--AI collaborative prompt loop, and reports extraction quality on a three-chapter gold set \cite{yuan2026knowledge}; agent-driven construction for Pre-Qin canonical texts organizes time and location as explicit entity types with a graph query answering layer \cite{zhang2026preqin}; and graph-driven spatial reconstruction of the Grand View Garden demonstrates what spatially indexed knowledge of this corpus enables \cite{wei2025daguanyuan}. However, these efforts extract without constraint checking, store the result as a static structure, and evaluate on ad hoc subsets rather than a released benchmark; to our knowledge no multi-hop QA golden set exists for classical Chinese literature, within the venue scope surveyed above, which leaves the quality of graph-grounded question answering on this corpus unmeasurable across systems. Our benchmark and our constraint-checked extraction address exactly this gap.

\subsection{Neuro-Symbolic Checking, Hallucination Control, and Temporal Graph Modeling}
Three adjacent technical lineages inform our mechanism design. Neuro-symbolic systems pair LLM reasoning with deterministic symbolic solvers and report faithfulness gains over neural-only pipelines \cite{pan2023logiclm}, and a systematic review of knowledge-graph integration for hallucination mitigation confirms that grounding output in verifiable triples is a widely studied strategy \cite{wagner2025mitigating}; the hallucination literature supplies both the taxonomy of failure modes and the atomic factual-precision metrics we adopt \cite{huang2024surveyhalluc,min2023factscore}. The complementary observation from that literature is that evaluation is the weak link: most hallucination benchmarks target generation rather than extraction, which is why our experimental design builds a dedicated extraction-level protocol instead of borrowing one. On the temporal side, rule-based temporal logic programs reason over temporal knowledge graphs with interpretable rules \cite{liu2022tlogic}, interval algebra underpins temporal consistency checking \cite{allen1983maintaining}, and the temporal graph embedding literature provides time-aware representations \cite{cai2024tkg,li2021regen}; spatio-temporal knowledge graph surveys show that unified modeling frameworks combining both dimensions remain largely absent, with most models tailored to single use cases \cite{plamper2025stkg}. Static embedding over small sparse knowledge graphs has also been studied, as a representative of the static alternative we depart from \cite{shi2023relagraph}. Narrative-time recovery from prose has its own lineage: narrative event chains induce partial event orderings from text \cite{chambers2008narrative}, and later work extends temporal ordering to richer document structures. That lineage recovers orderings but stops at the timeline: it does not feed a constraint-checked knowledge graph, does not attach validity intervals to extracted relations, and does not support slice retrieval for question answering. What the lineages do not yet combine is therefore narrower and more precise than a general absence claim: a constraint-checked extraction layer feeding a temporally dynamic graph with slice retrieval; rule checking exists for reasoning over an already-built graph, temporal modeling exists for data whose time stamps are given, and narrative ordering exists as a standalone prediction task, but no system connects narrative-time recovery to extraction-time constraint checking and temporal-slice QA in one pipeline.

The single nearest prior work is the human--AI collaborative construction for our source novel \cite{yuan2026knowledge}: it shares the corpus and the LLM-driven extraction setting, but it applies no ontology constraint layer, builds a static triple set, offers no temporal or spatial coordinates, and releases no QA benchmark. Its extraction loop improves yield through prompt iteration, whereas our constraint layer addresses the complementary problem: unrestricted iteration can grow the triple set without growing its reliability, and reliability is exactly what a constraint-checked, evidence-anchored extraction is designed to guarantee. Our framework occupies precisely the space it leaves open, which the methodology section now specifies.

\section{Methodology}
\subsection{Problem Formulation}
We consider knowledge processing over long-form literary narrative: a corpus $D = \{c_1, \ldots, c_K\}$ of $K$ chapters, where each chapter $c_k$ is a continuous prose segment of several thousand characters.

\begin{figure}[ht]\centering
\includegraphics[width=\textwidth]{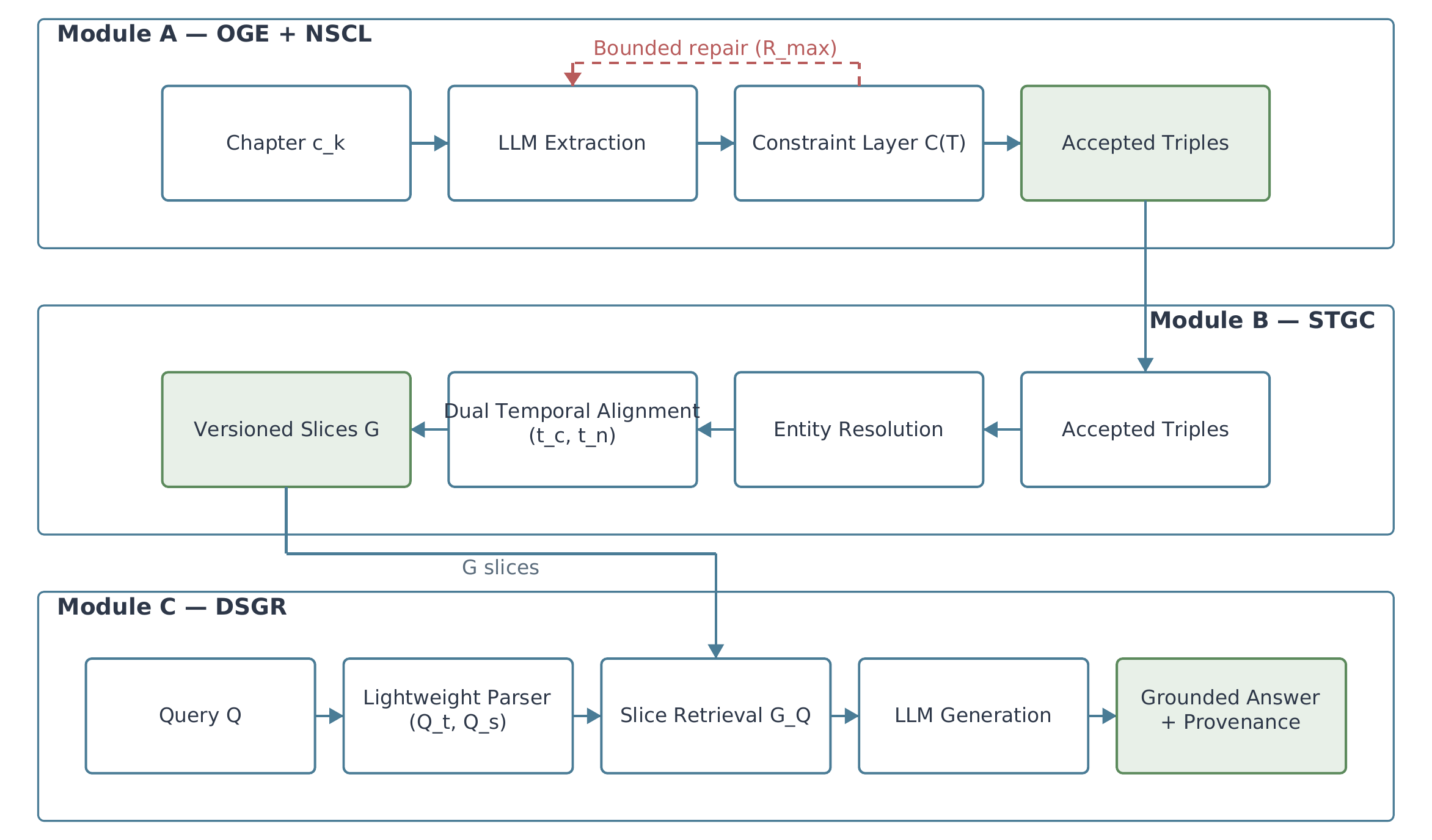}
\caption{Overview of the NS-ST-GraphRAG pipeline, organized into three swimlanes. Module A (OGE + NSCL): a chapter $c_k$ enters LLM extraction; candidate triples pass through the constraint layer $C(\mathcal{T})$; the dashed feedback arrow marks the bounded repair loop, which re-prompts extraction on violating spans at most $R_{max}$ times, and accepted triples leave the module. Module B (STGC): accepted triples undergo entity resolution, receive dual temporal coordinates (chapter index $t_c$ and narrative time $t_n$), and are stored as versioned slices $G$. Module C (DSGR): a query $Q$ is parsed into temporal and spatial constraints $Q_t, Q_s$ by a lightweight parser; the matching slice $G_Q$ is retrieved by index lookup; LLM generation produces the grounded answer with provenance. Solid arrows carry data flow; dashed arrows carry feedback.}\label{fig:system_overview}
\end{figure}

The target of construction is a knowledge graph $G = (V, E)$ whose entities $V$ are characters, locations, objects, and events, and whose relations $E$ include kinship, master-servant bonds, social interactions, and event participation. Two properties distinguish this setting from conventional knowledge graph construction \cite{ji2021survey}. First, the narrative is temporally organized: relations hold within specific narrative intervals and change as the story unfolds, so a static triple set loses information that the text explicitly carries. Second, the prose is dense with metaphor, allusion, and indirect reference, which makes unrestricted large-language-model extraction prone to producing relations that contradict the story's internal logic, such as impossible kinship links or master-servant bonds across rival households. The two difficulties interact: temporal modeling errors and hallucinated relations compound, because a relation asserted at the wrong narrative moment is as misleading as a relation that never existed. Our objective is a pipeline that extracts a spatio-temporal graph with a low rate of such hallucinated relations, supports queries constrained by time and place, and grounds every generated answer in triples whose narrative coordinates can be traced. This objective is stated formally in the notation below and realized by the three modules that follow. Figure~\ref{fig:system_overview} summarizes the pipeline.

\subsection{Notation}
Table~\ref{tab:notation} fixes the symbols used throughout. The graph is indexed by two temporal coordinates, a coarse chapter index and a fine narrative-time interval, and by a spatial coordinate drawn from the set of scenes and locations mentioned in the text.

\par\noindent\begin{minipage}{\columnwidth}\centering
\captionsetup{type=table}
\captionof{table}{Notation used throughout the paper.}\label{tab:notation}
\begin{tabularx}{\columnwidth}{@{}l X@{}}\toprule
Symbol & Meaning \\ \midrule
$G=(V,E)$ & spatio-temporal knowledge graph with entity set $V$ and relation set $E$ \\
$O$ & domain ontology rule set (genealogical, master-servant, title and gender constraints) \\
$C(\cdot)$ & hard-rule filtering function over candidate triples \\
$P(\cdot), \theta$ & soft-constraint penalty function and its threshold \\
$\tau(e)$ & temporal coordinate interval of entity or edge $e$ \\
$\sigma(e)$ & spatial coordinate (scene or location) of entity $e$ \\
$t_c$ & chapter index, the coarse temporal coordinate \\
$t_n$ & narrative-time event interval, the fine temporal coordinate \\
$Q_t, Q_s$ & temporal and spatial constraints parsed from query $Q$ \\
$G_Q$ & temporal slice of $G$ retrieved for query $Q$ \\
$R, R_{max}$ & repair round count and its upper bound \\
\bottomrule
\end{tabularx}

\end{minipage}\par

\subsection{Ontology-Guided Extraction with a Neuro-Symbolic Constraint Layer}
The first module, Ontology-Guided Extraction (OGE), produces candidate triples under an explicit domain ontology, then filters and repairs them through the neuro-symbolic constraint layer (NSCL). Extraction proceeds chapter by chapter: each chapter segment is passed to an LLM with a schema prompt that names the ontology's relation types and requests, for every triple, the supporting evidence span and the chapter in which it holds. This schema-constrained prompting is the neural half of the layer; the symbolic half operates on its output.

The ontology $O$ encodes three structural families of rules that capture the structural logic of the narrative domain.

\begin{figure}[ht]\centering
\includegraphics[width=\textwidth]{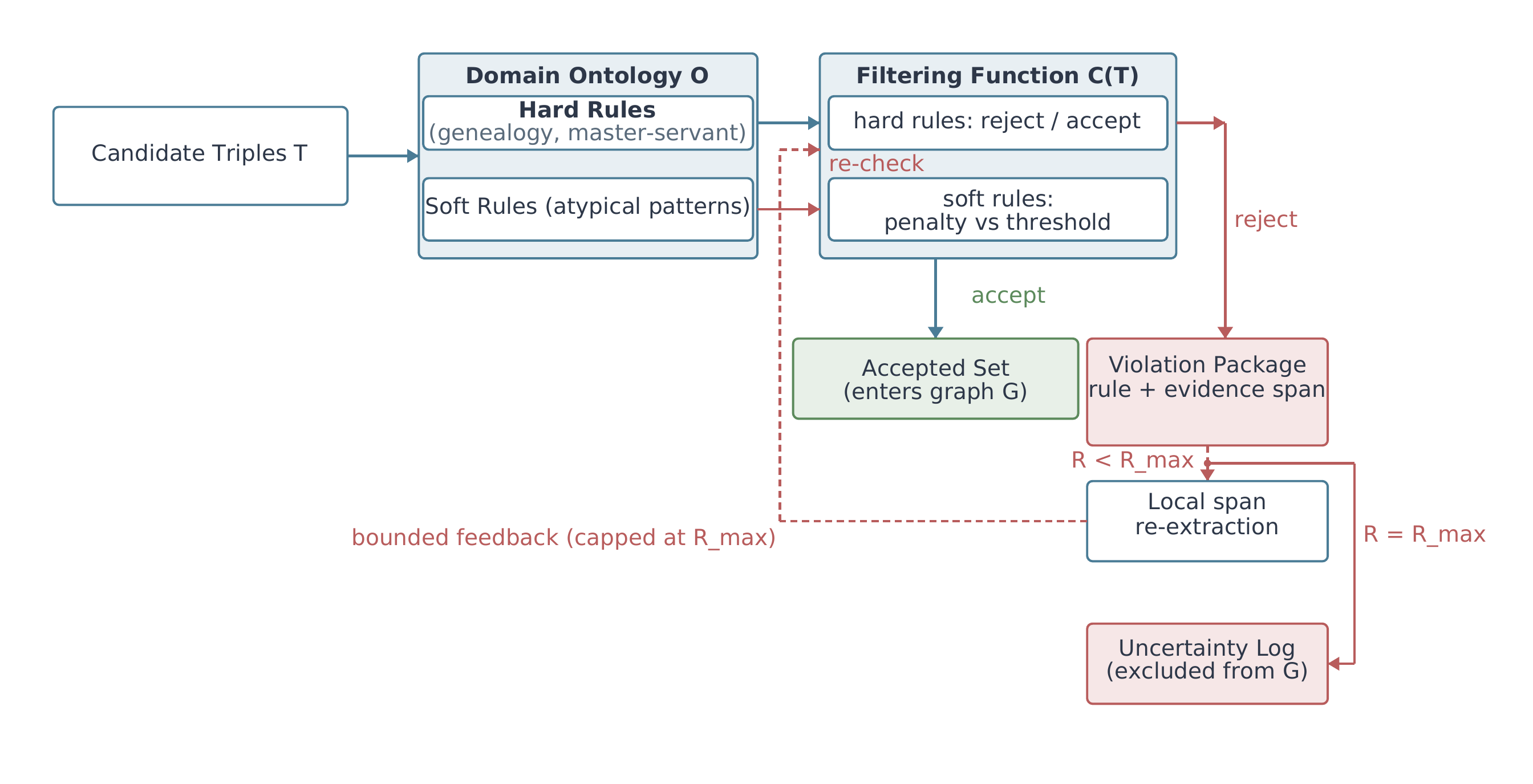}
\caption{The neuro-symbolic constraint layer (NSCL). Candidate triples $\mathcal{T}$ from extraction are evaluated against the domain ontology $\mathcal{O}$, which branches into hard rules (genealogical, master-servant, and title/gender consistency) and soft rules (atypical patterns). The filtering function $C(\mathcal{T})$ applies a binary reject/accept decision for hard rules and a penalty score against a fixed threshold for soft rules. Hard-rule rejections form a violation package carrying the offending rule and the evidence span, which feeds the dashed feedback loop back to extraction, capped at $R_{max}$ repair passes; hard violations still present after the cap are written to the uncertainty log and excluded from the graph $G$. Soft-rule penalties above the threshold mark the triple with a low-confidence attribute instead of excluding it (not depicted).}\label{fig:constraint_layer}
\end{figure}

Figure~\ref{fig:constraint_layer} details the layer. Genealogical constraints enforce consistency of kinship: parenthood is closed under consistency checks that forbid conflicting parent assignments and detect impossible cycles within a lineage. Master-servant mutual exclusion forbids a servant bound to one household from simultaneously appearing as a close consanguineous relative of a rival household, a pattern that the source text never licenses but that LLMs readily hallucinate under co-occurrence pressure. Title and gender constraints check that extracted attributes agree with the honorifics and address terms used in the evidence spans. Each rule is expressed as a small first-order pattern over the triple vocabulary, and the constraint layer applies them as a deterministic filtering function
\begin{equation}
C(\mathcal{T}) = \bigl\{ r \in \mathcal{T} : r \models O \bigr\},
\end{equation}
where $\mathcal{T}$ is the candidate triple set of a chapter and $r \models O$ denotes satisfaction of all applicable hard rules. Violated triples are not silently discarded: each hard violation is packaged with the offending rule and the evidence span, and the LLM is re-prompted to repair the extraction on that span only. This feedback loop is bounded at $R_{max}$ rounds, with $R_{max}$ set to a small fixed value in the released configuration; triples that still violate a hard rule after the bound are excluded from $G$ and logged with an uncertainty attribute. The chapter-local filtering is followed by one post-assembly global pass over $G$, which evaluates the rules whose scope spans chapters, such as kinship-cycle detection and conflicting parent assignments, by transitive closure over the assembled graph; the pass runs once per build at cost linear in the number of kinship edges and is included in the construction-cost accounting of the experimental design. Soft constraints, such as unusual but not impossible co-occurrence patterns, contribute a weighted penalty instead of a hard rejection: the penalty weights are fixed by the domain expert at ontology authoring time, the penalties of a triple are summed over the violated soft rules, and a triple is admitted when its total penalty stays below a fixed threshold, otherwise it is kept in $G$ with a low-confidence attribute rather than rejected. Soft violations never enter the repair loop or the exclusion path; only hard-rule violations are repaired and, after the cap, excluded. This keeps the boundary between impossible and merely atypical relations explicit while the admission rule itself remains a deterministic, auditable computation.

The ontology itself is acquired through an LLM-assisted, expert-verified procedure rather than hand-authored from scratch. Rule candidates are generated by mining extraction errors on a development split of the corpus, in which a first-pass extraction without constraints is compared against expert annotations and the recurring error patterns are summarized into rule proposals. A domain expert then verifies each proposal, and the paper reports the resulting rule count and the verification effort in person-hours. We treat this acquisition cost as a reported, auditable quantity rather than an unexamined assumption: the protocol bounds it by construction, because the candidate generation is automated and the expert's role is verification, not invention. Auditability is likewise a designed property rather than a side effect. Every rejection is reproducible: it names the rule that fired, and any reviewer can re-run that rule over the released ontology and confirm the outcome. This converts hallucination control from an asserted property of a prompt into an inspectable mechanism of a system, which is the standard a measurement-driven evaluation requires.

The soft-constraint branch deserves a precise statement, because the hard-soft distinction is where the layer's conservatism is controlled. A soft constraint is implemented as a weighted penalty term over the triple's evidence features, so an atypical but textually supported relation survives while an unsupported one is discouraged; the weights are fixed by the expert at ontology authoring time and held constant across corpora in the evaluation. Evidence-span quality is enforced jointly with the rules, and its test is operationalized deterministically: a triple whose supporting span shares no surface form with any alias of either entity is treated as a soft violation, which penalizes the extraction style in which an LLM asserts a plausible relation without pointing to the passage that licenses it; the check is a string-level alias match with no model call. This coupling between ontology rules and evidence anchoring is what makes the provenance metric of the experimental design meaningful, since a relation that cannot be anchored cannot later ground an answer.

We contrast this design with two alternatives in the design rationale below. A pure post-hoc LLM checker, in which a second model pass judges triples without explicit rules, is not deterministic, cannot be audited rule by rule, and costs one additional LLM call per triple. A fully learned verifier would require training data for a phenomenon we explicitly want to measure, since the gold annotation for hallucination is the evaluation target itself. Rule-based checking is deterministic and auditable, which motivates its position as the backstop layer over neural extraction \cite{pan2023logiclm,wagner2025mitigating}.

Algorithm~\ref{alg:oge} states the full extraction procedure.

\begin{algorithm}[htb]
\caption{Ontology-guided extraction with constraint repair.}\label{alg:oge}
\begin{algorithmic}[1]
\Require chapters $\{c_1,\dots,c_K\}$, ontology $O$, round bound $R_{max}$, soft penalty $P$ with threshold $\theta$
\Ensure candidate triples $\mathcal{T}$, accepted set $\mathcal{A}$, uncertain log $\mathcal{U}$, low-confidence set $\mathcal{L}$
\State $\mathcal{A} \gets \emptyset$; $\mathcal{U} \gets \emptyset$; $\mathcal{L} \gets \emptyset$
\For{each chapter $c_k$}
  \State $\mathcal{T}_k \gets \textsc{Extract}(c_k)$ \Comment{schema-prompted LLM pass}
  \State $r \gets 0$
  \While{$\mathcal{T}_k \setminus C(\mathcal{T}_k) \neq \emptyset$ and $r < R_{max}$}
    \State $\mathcal{V} \gets \mathcal{T}_k \setminus C(\mathcal{T}_k)$ \Comment{hard violations with rule + span}
    \State $\mathcal{T}_k \gets \textsc{Repair}(\mathcal{T}_k, \mathcal{V})$ \Comment{span-local re-extraction}
    \State $r \gets r + 1$
  \EndWhile
  \State $\mathcal{A}_k \gets C(\mathcal{T}_k)$
  \State $\mathcal{A} \gets \mathcal{A} \cup \mathcal{A}_k$
  \State $\mathcal{U} \gets \mathcal{U} \cup (\mathcal{T}_k \setminus C(\mathcal{T}_k))$
  \State $\mathcal{L} \gets \mathcal{L} \cup \{\, t \in \mathcal{A}_k : P(t) \geq \theta \,\}$ \Comment{soft violations kept with a low-confidence attribute}
\EndFor
\State \Return $\mathcal{A}, \mathcal{U}, \mathcal{L}$
\end{algorithmic}
\end{algorithm}

\subsection{Spatio-Temporal Graph Construction}
The second module, Spatio-Temporal Graph Construction (STGC), assembles the accepted triples into a graph whose entities and edges carry temporal and spatial coordinates.

\begin{figure}[ht]\centering
\includegraphics[width=\textwidth]{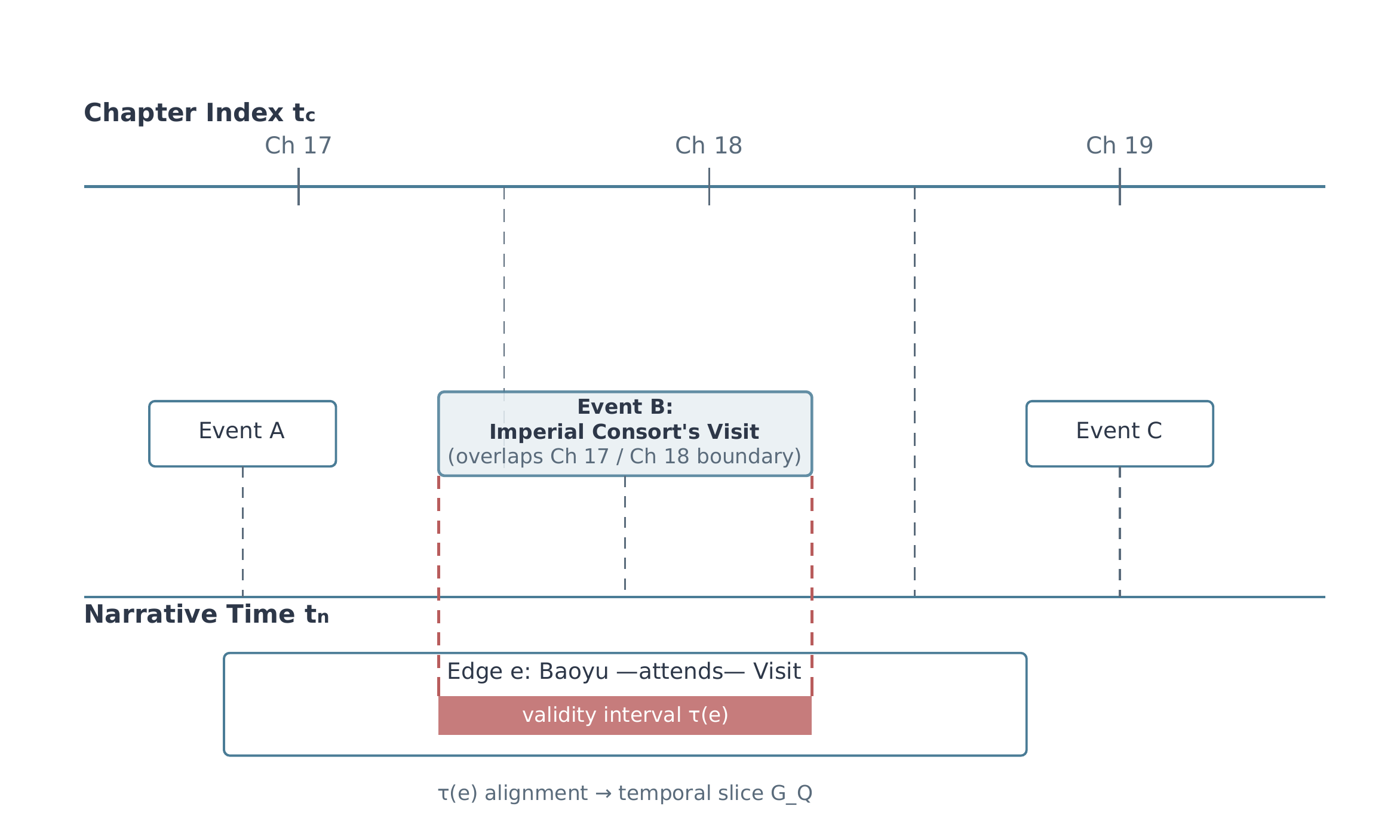}
\caption{Dual temporal coordinate system. The top axis is the coarse chapter index $t_c$ with chapter markers (Ch 17, Ch 18, Ch 19); the dashed vertical lines mark chapter boundaries. The bottom axis is the fine narrative time $t_n$ with story events as intervals: Event B (the Imperial Consort's Visit) is drawn as a semi-transparent block deliberately overlapping the Ch 17 / Ch 18 boundary, illustrating why chapter indices alone cannot locate events. Below the axes, a knowledge graph edge $e$ (Baoyu attends the Visit) carries a validity interval $\tau(e)$ rendered as the red band; the dashed connectors anchor its endpoints to the narrative timeline so that $\tau(e)$ aligns exactly with Event B's extent. This alignment is the mechanism by which time-constrained queries select their slice $G_Q$.}\label{fig:temporal_model}
\end{figure}

Chapter indices alone are too coarse for narrative modeling: a single chapter may cover several days of story time, while major events, such as the Imperial Consort's visit in the source corpus, span multiple chapters. We therefore maintain a dual temporal coordinate system, illustrated in Figure~\ref{fig:temporal_model}. The coarse coordinate is the chapter index $t_c$. The fine coordinate is the narrative time $t_n$: an LLM event-timeline pass over the corpus extracts story events with their temporal extents, and every edge receives a validity interval $\tau(e)$ over this narrative timeline. The value domain of $t_n$ is defined concretely: the timeline pass emits a global set of anchor events, chosen as events whose relative order is explicit in the text, and every extracted interval is expressed as a pair of anchor endpoints with a relation from a tractable subset of Allen's interval relations \cite{allen1983maintaining} (the five relations needed to express before, after, overlapping, containing, and contained), checked pair-wise over adjacent anchors rather than by global consistency search, which keeps the check linear in the number of intervals. The ontology $O$ carries a fourth rule family, temporal consistency rules, that ties the two coordinate systems together: the validity interval of an edge extracted from chapter $c_k$ must fall within the interval of $c_k$, and an interval that overlaps two events the timeline orders as disjoint is flagged and repaired through the same bounded mechanism as extraction errors. Spatial coordinates $\sigma(e)$ assign entities to scenes and locations mentioned in their evidence spans, which supports the space-constrained query category defined in the experimental design.

Entity resolution precedes graph assembly. Literary texts refer to a character by many aliases and honorifics; in the source corpus a single protagonist appears under names such as Fengjie, Sister Feng, and the Second Mistress Lian. Resolution proceeds by LLM alias unification, in which candidate name clusters are proposed from co-occurrence and address patterns. Each merge is validated by two independent signals: the constraint layer, so that two aliases attached to conflicting genealogical positions cannot be merged, and evidence-span co-occurrence support, which requires the aliases to appear in at least one shared scene context before a merge is accepted. The merge error rate is measured on the annotation subset in the experimental design. Detecting and disambiguating characters in literary prose is a known-hard task \cite{vala2015mrbennet}, and the ontology acts as the tie-breaker that prior work on network extraction lacked.

The event-timeline pass that recovers narrative time has its own internal discipline. The LLM is prompted to extract story events with their temporal extents relative to anchor events whose order is explicit in the text, and to place each extracted relation's evidence span within that event structure rather than against calendar dates, which the source corpus rarely states. In the full design, after resolution the graph is stored in a snapshot-plus-delta model: a materialized snapshot slice after every chapter, plus event-delta slices that record the changes a major event introduces. The snapshot per chapter makes storage linear in the number of relations times the number of chapters, and the deltas keep event-level queries from requiring reconstruction from scratch; a query slice $G_Q$ is reconstructed as the latest snapshot at or before the query's narrative point, overlaid with the event deltas whose intervals fall inside the query window, at cost linear in the size of the returned slice plus a logarithmic lookup. This storage design also supports the change-over-time question class directly: a before-and-after query retrieves the two boundary snapshots and returns their difference, which is the operation the evolution case study of the analysis section builds on.

The design rationale follows from a direct contrast with static graph construction. GraphRAG-style pipelines build a corpus-level entity-relation graph without explicit narrative-time validity intervals and answer queries against it \cite{edge2024graphrag,guo2024lightrag,gutierrez2024hipporag}; for a question about the state of affairs before and after a narrative turning point, the graph itself provides no explicit mechanism for scoping retrieval to one side. Static embeddings over knowledge graphs face the same boundary \cite{shi2023relagraph}. Temporal knowledge graph models provide time-aware representations \cite{trivedi2017knowevolve,sadeghi2021chronor,li2021regen}, and temporal graph benchmarks standardize their evaluation \cite{huang2023tgb}; these models and benchmarks generally operate on temporal information already provided in the data, whereas narrative time must first be recovered from prose. Our dual coordinate system recovers narrative time with an explicit extraction pass and keeps the chapter index as a fallback that never requires the timeline pass to be perfect.

The question of whether a graph is needed at all deserves an explicit answer, because a reviewer may reasonably ask why a long-context LLM should not answer directly over the corpus. Three information processing properties separate the graph route from the direct route. Retrieval efficiency: a direct model pays the full context-window cost of the relevant chapters on every query, whereas slice retrieval costs a bounded lookup plus the slice, so query cost scales with the question's temporal scope rather than with corpus size. Context limits: the full corpus exceeds practical context windows, and any chunking reintroduces the very routing problem the temporal index solves. Structural auditability: a direct model's answer cannot be decomposed into triples with narrative coordinates, so the provenance metric that anchors the evaluation has nothing to inspect, and hallucination control reduces to trust in the model. The graph is not merely a representational convenience; it is the artifact that makes retrieval efficiency measurable, provenance checkable, and grounded generation mechanically constrained to evidence spans.

\subsection{Dynamic Sub-Graph Retrieval}
\begin{figure}[ht]\centering
\includegraphics[width=\textwidth]{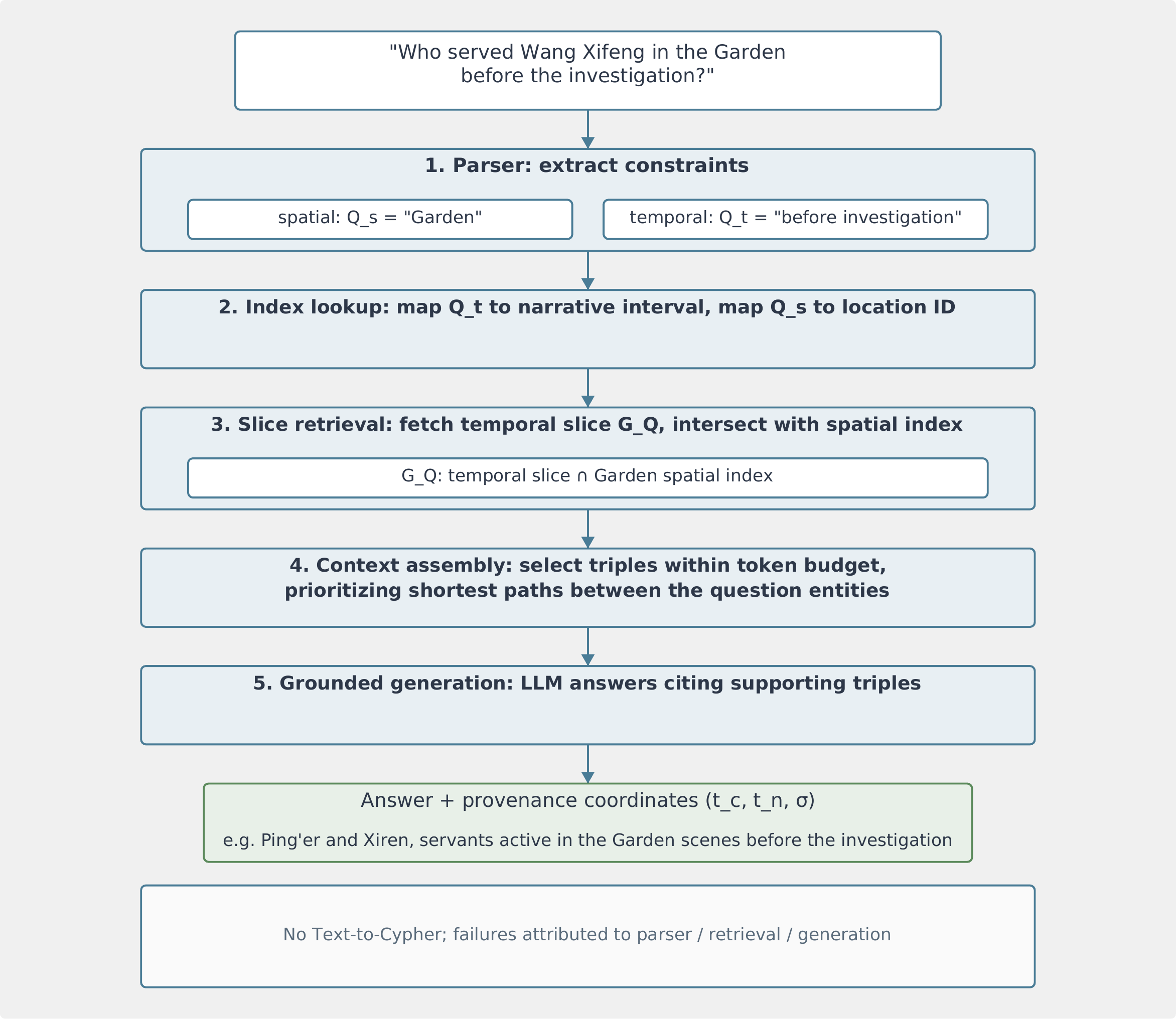}
\caption{A time-constrained query traced through Module C using the running example ``Who served Wang Xifeng in the Garden before the investigation?'' (1) The parser extracts a spatial constraint $Q_s$ = Garden and a temporal constraint $Q_t$ = before the investigation. (2) Index lookup maps $Q_t$ to a narrative-time interval and $Q_s$ to a location ID. (3) Slice retrieval fetches the temporal slice $G_Q$ and intersects it with the scene-to-entity spatial index. (4) Context assembly selects triples within the token budget, prioritizing shortest paths between the question's entities so multi-hop chains survive truncation. (5) Grounded generation answers from the selected triples and returns provenance coordinates $(t_c, t_n, \sigma)$. No graph query language statement (e.g., Cypher) is generated at any step (illustrative trace, no numeric results).}\label{fig:query_routing}
\end{figure}

The third module, Dynamic Sub-Graph Retrieval (DSGR), answers queries under temporal and spatial constraints, traced in Figure~\ref{fig:query_routing}. Parsing is deliberately lightweight. An intent-recognition pass extracts the constraint entities from the query, a temporal condition such as a chapter range or a story event, and a spatial condition such as a scene, and links them to the graph's temporal and spatial identifiers. Retrieval then selects the matching slice $G_Q$ by structured index lookups over the versioned slices, filtered by a scene-to-entity spatial index built alongside the graph, so a space-constrained question intersects its slice with the entities of the named scenes; generation proceeds from the selected triples with their evidence spans in context. We do not generate a graph query language statement, such as Cypher, from the user question, and the rejection is a measurement decision rather than a preference. A text-to-query stage is an additional learned component with its own uncontrolled error rate: its failures would be indistinguishable from retrieval failures in the evaluation, directly confounding the effectiveness attribution of RQ2, and its latency adds an uncontrolled term to the retrieval-efficiency measurement of RQ3. Identifier linking plus index lookup contains both risks. Its errors are attributable, because the parser metric isolates them before retrieval is reached, and its cost is a bounded index lookup rather than a generation pass, so the RQ3 comparison measures the temporal-slice mechanism itself and nothing else.

Grounded generation closes the loop with provenance. Every answer is required to map onto a set of supporting triples, and each triple carries its chapter, narrative-time, and scene coordinates, so the answer's basis can be traced and checked. The fraction of answers whose supporting triples are verifiable defines the provenance tracing accuracy metric of the experimental design. The query-time token budget is a design constraint in its own right: the slice's triples plus their evidence spans are placed in context under a budget, and when a slice is large, the retriever ranks triples by graph distance from the constraint entities in the slice, measured over the relation graph, with ties broken by evidence recency relative to the query's narrative point. To protect multi-hop chains from truncation, the shortest paths between the question's own entities are always retained before any distance-based pruning, so a hop that is far from the constraint entities but on the answer path survives the budget cut. Constraint parsing is measured separately on the time- and space-constrained question subset, and question failures are attributed to one of three stages, parsing, retrieval, or generation, so that the effectiveness and efficiency claims of the temporal-slice design are not confounded by parser noise.

\subsection{Construction-Cost Structure}
The pipeline's construction cost is a design object rather than an accident of implementation. The token budget of a full build decomposes into the schema-prompted extraction pass and the event-timeline pass, both linear in the corpus size, plus the repair rounds, bounded by $R_{max}$ per chapter. The ontology checks themselves, including the post-assembly global pass, require only deterministic rule evaluations and no additional model calls. The repair loop re-extracts only the violating spans rather than whole chapters, so its cost scales with the violation rate times the corpus size, with the multiplier bounded by the rounds-until-success capped at $R_{max}$; a corpus on which the schema prompt performs well incurs repair cost only on the small subset of violating spans. We compare this budget against the indexing cost of the baseline systems in the experimental design, and the ablation isolates the incremental cost of the constraint layer, which supports an explicit cost-benefit argument: the layer is justified if its hallucination reduction is obtained at a construction-cost premium that the ablation shows to be small relative to the base extraction pass.

\section{Experimental Results}
\subsection{Implementation Details}
The pipeline is implemented with two LLM passes driven by DeepSeek-V3.2 at temperature $0$ (the answering model of the baselines) accessed through its API---schema-prompted extraction and the event-timeline pass, one call per chapter each in the base build---and the constraint layer, chapter-coordinate repair, entity resolution, graph assembly, and retrieval indexes implemented as deterministic Python components with no model calls. The ontology rules are stored as small first-order patterns over the triple vocabulary; the repair loop is bounded at $R_{max}$ rounds per chapter and the shipped full-corpus build runs it at $R_{max}=0$, deterministic repair only. The event-timeline pass emits event occurrences with (chapter, sequence) narrative coordinates and anchor-event references, indexed by chapter for slice lookup. The versioned slices are kept in memory for the primary corpus, and the slice index is a chapter-to-slice mapping with event-range lookup; the snapshot-plus-delta storage model of the design is deferred to the full build. The full configuration, prompts, and seeds are archived for release with the implementation repository upon acceptance.

\subsection{Experimental Design}
The primary corpus is Dream of the Red Chamber, all 120 chapters, in the public-domain original text (the ctext.org digitization of the Cheng-Yiben lineage, the same chapter files released with the benchmark). The generalization corpus is Camel Xiangzi (Rickshaw Boy), a modern vernacular Chinese novel whose author died in 1966, placing the Chinese original in the public domain in life-plus-fifty jurisdictions since 2017; the benchmark release ships acquisition scripts and a canonical source pointer---the standard People's Literature Publishing House edition, with a public-domain digitization (public in life-plus-fifty jurisdictions since 2017) as the acquisition target---rather than redistributing the text, since the work remains protected in life-plus-seventy jurisdictions until the end of 2036. Its Republican-era Beijing setting supplies the urban scene structure and its multi-year plot the narrative timeline, and it tests cross-genre, same-language transfer from the classical household novel rather than a second classical text, which would only probe genre-internal variation. The ontology acquisition cost---rule count and expert person-hours---is tracked on each corpus as a first-class result, so the transfer cost of the symbolic layer is visible in the same evaluation as its benefit.

A first sampling measurement of the generalization corpus is reported here rather than deferred. Three chapters spanning the narrative (1, 12, 23) were extracted with the same schema prompt, substituting only the book's title; the 31 raw items, judged independently by two annotators and reconciled item by item, reduce to 26 unique relations after deduplication. Under the primary corpus's unified judgment criteria, two violations remain on this baseline ($7.7\%$ of the 26 unique relations): one non-existent relation and one wrong-level kinship predicate; the raw round had judged the kinship item as needing revision rather than a violation, so the raw 31-item strict rate is $3.2\%$ ($1/31$)---the reclassification is the cross-corpus criterion unification, recorded per item in the release. The pipeline's deterministic repair transfers intact---the copied-coordinate failure replicates on all 27 items of Chapters 12 and 23 (the four Chapter-1 items carry the copied value correctly) and is fixed mechanically, consistent with a prompt-induced rather than corpus-specific failure. The adaptation phase then applied the design's bounded-cost clause: of the 26 deduplicated items, six were fixed by evidence or predicate adjustment, and seven were deleted ($26.9\%$ attrition against the strict-evidence standard). Additionally, one predicate was added to the schema vocabulary to accommodate modern vernacular possession. The remaining 19 items (6 fixed plus 13 retained as-is) all pass validation, yielding a post-adaptation strict hallucination rate of $0\%$ at an expert-estimated adaptation cost of 25--35 minutes. The modern vernacular quotes more faithfully ($1$ of $31$ items non-verbatim against $6$ of $98$ on the classical sample---exploratory, across non-comparable samples) while the classical rule families stay nearly silent (one type-signature flag). The sample is small and the cost figure is an expert's estimate over an auditable operation list, not a timed measurement; the operation list and the two annotators' item-level judgments ship with the release, and the full-corpus generalization measurement remains scheduled.

Four baselines are compared under a prompt-equalization protocol: Naive Vector RAG over chapter chunks; GraphRAG with community summarization \cite{edge2024graphrag}; LightRAG with dual-level retrieval \cite{guo2024lightrag}; and HippoRAG with personalized PageRank retrieval \cite{gutierrez2024hipporag}. The set is the graph-grounded retrieval family plus the vector floor. Every baseline receives the same domain background and few-shot examples as the proposed system, and equal tuning effort under a fixed prompt-iteration budget, recorded and reported; the ontology is part of the architecture under test, not an external resource granted to one competitor. To keep the temporal dimension testable, one additional condition is evaluated: a chapter-scoped Naive RAG variant whose retrieval window is restricted to the gold chapter range, giving the static systems a temporal-capable counterpart at the cost of leaking the gold range, a leak the proposed system does not receive.

The metrics are extraction entity F1 and relation F1; multi-hop QA accuracy overall and by category (time-constrained, space-constrained, general); hallucination rate, split into extraction-level (fraction of extracted triples violating golden annotation) and grounding-level (fraction of answer-grounding steps violating golden annotation), both measured on an annotation subset disjoint from rule tuning; temporal-coordinate accuracy, scored by interval overlap between extracted validity intervals and gold narrative-time annotations; provenance tracing accuracy, the fraction of answers whose cited triples carry verifiable chapter, event, and scene coordinates, judged by deterministic span matching; retrieval latency with the generation call excluded; and indexing construction cost in tokens, API cost, and wall-clock time. Multi-hop QA accuracy differences are tested with McNemar's test and bootstrap intervals, with pairwise system-by-category comparisons pre-registered and Holm-Bonferroni adjusted. The frozen protocol and its revision history ship with the release. Questions are verified to require multiple evidence pieces rather than lexical shortcuts \cite{yang2018hotpotqa,ho2020constructing,chen2019understanding}, answer correctness is judged against the golden set with atomic-fact precision following \cite{min2023factscore}, the failure-category analysis follows the categorization and evaluation design of CRAG \cite{yang2024crag}, and the evolution-query design is informed by temporal graph benchmarks \cite{huang2023tgb}.

The extraction and hallucination gold is specified in advance and released with the benchmark: a fixed set of chapters spanning the early, middle, and late narrative is annotated at triple granularity, with its size and annotation cost reported with the full campaign, so the paper's own criticism of ad hoc gold sets does not apply to itself. Chapter-level disjointness is enforced across all annotation layers (rule-candidate mining, timeline tuning, hallucination annotation, and the QA test split); the per-pair chapter overlaps are stated in the release and reproduced by its scripts, and the trajectory section below flags explicitly that the frozen R5 configuration is a development-set maximum, not a test-set estimate.

The test split is used exactly once; rule tuning, prompt selection, and threshold choices happen on train and validation splits only, with prompts frozen for the final run. The preliminary released set serves mechanism and protocol development; the full campaign's test split is disjoint from it and used exactly once under the frozen configuration, as reported in the Main Campaign subsection. A human expert baseline is pre-registered to anchor difficulty: experts who did not annotate the gold set will answer open-book under a stated time budget, with accuracy, agreement, and person-hour cost reserved for the complete release.

Question candidates are generated LLM-assisted from chapter evidence and verified by a domain expert, who rewrites questions that admit shortcuts, confirms the supporting evidence set for each answer, and assigns the category. Time-constrained answers carry narrative-time coordinates rather than chapter numbers alone, keeping the evaluation aligned with the dual-coordinate model. Inter-annotator agreement is measured on a verification sample and reported with the benchmark release, which includes the evaluation scripts, so the numbers reported in Section~4 are reproducible by third parties.

\subsection{Preliminary Results}

This revision reports the first measured installment: the preliminary release of Red-Chamber-QA with three answer-generation baselines and a mechanical ceiling. The graph-augmented baselines, the full system comparison, the extraction-quality and latency metrics, the full-corpus generalization measurement, and the ablations remain part of the pre-registered design and are scheduled for the complete release; the campaign hypotheses of Section~5 are not tested here. Every number below traces to the released benchmark and evaluation scripts (the 104-question preliminary set is public; the implementation repository is withheld during review and linked on acceptance; the frozen protocol and its revision history ship with the release).

Preliminary benchmark set. The released set contains 104 questions over 60 chapters: each carries a verbatim premise fragment, a two-part golden answer, per-part evidence spans, and a category label, passed the mechanical construction rule set---answer-in-stem overlap, sub-eight-character premise anchors, insufficient premise-answer distance, and premise-anchor order contradictions are rejected mechanically---and a domain expert rewrote any remaining shortcut-admitting question, so the $0\%$ shortcut rate ($0/104$) holds by construction; the rule set ships with the benchmark scripts and its four checks are deterministic and re-runnable. The distribution is General $81$, Time-Constrained $15$, Space-Constrained $8$; $88$ single-hop and $16$ multi-hop; $63$ hard and $41$ medium under the release's difficulty rubric.

Baselines and protocol. Three baselines are evaluated: a closed-book LLM with no source text (the same prompt with an empty context), a long-context LLM receiving the full required chapters, and a naive chunk retrieval system (character-bigram BM25-lite) whose top-5 chapter-block passages feed the same LLM, where blocks are fixed-width character windows of $2000$ characters segmented at chapter boundaries (the source digitization carries no paragraph marks); the pre-registered family lists Naive Vector RAG over chapter chunks; this installment substitutes a dependency-free BM25-lite retriever, which the full campaign will reconcile by reporting the vector baseline as pre-registered. An oracle baseline mechanically emits golden answers with evidence spans as a metric-logic ceiling. All answering-model calls use DeepSeek-V3.2 at temperature $0$, with two frozen few-shot demonstrations from chapters disjoint from every split; only the judge uses a second, independent model (Qwen3.5-35B-A3B). Answer quality is scored on two tracks of dual-part completeness (DPC): DPC-mech counts a two-part answer correct only if each part is reproduced mechanically---golden answer fragment or verbatim evidence span, exact or with an overlap of at least eight characters; the evidence span is the verbatim channel because the golden fragments are expert-phrased paraphrases; DPC-judge has an independent cross-model judge decide each part open-book against the full chapter text, validated on a 10-question pilot disjoint from the scored set (agreement $0.80$--$0.90$, $\kappa$ $0.41$--$0.78$, zero gold misjudgments); the $\kappa$ interval is wide because $n=10$, and the full campaign reports per-question agreement with confidence intervals. The judge evaluates one answer against one golden part in a single-message format without positional ordering, so position bias is excluded by construction, and the answer's length is bounded by the cited evidence field the assembly emits, so verbosity is not otherwise rewarded. DPC-judge is computed over answers for which the judge returned a verdict after three retries; the excluded answers are reported in the table's valid-$n$ column (59/71/68/69). Counting no-verdicts as incorrect gives lower bounds $0.250$/$0.625$/$0.567$/$0.663$ (closed-book/long-context/naive RAG/oracle); the oracle's 35 no-verdicts are retry exhaustions, not misjudgments: the judge is a reasoning model whose thinking budget consumes the output allowance on long-evidence questions, producing empty verdict content, and the retries inherit the same failure mode---the cause is diagnosed, not assumed. Citation faithfulness (CF) is the deterministic verbatim-citation rate: the evidence passes only if every fragment of at least eight characters it contains is a verbatim substring of the required chapters---a mechanical verbatimness check, not a semantic relevance judgment. CF is scored against the gold chapter range rather than each system's provided context, so all systems face one citation standard; quoting outside the gold range is penalized accordingly---the intended strictness.

\begin{table}[htb]
\caption{Main results on the preliminary set (104 questions), with 95\% Wilson intervals. CF and DPC-judge are the citation and semantic-accuracy signals; DPC-mech is the strict near-literal anchor (exact match or an overlap of at least eight characters).}
\label{tab:main_results}
\centering\normalsize\setlength{\tabcolsep}{3pt}

\adjustbox{max width=\columnwidth}{%
\begin{tabular}{lcccc}\toprule
Method & CF & DPC-mech & DPC-judge & \begin{tabular}{@{}c@{}}DPC-judge\\ valid $n$\end{tabular} \\ \midrule
Closed-book LLM & $0.000$ [$0.000$,$0.036$] & $0.125$ [$0.075$,$0.202$] & $0.441$ [$0.322$,$0.567$] & 59 \\
Long-context LLM & $0.510$ [$0.415$,$0.604$] & $0.231$ [$0.160$,$0.320$] & $0.915$ [$0.828$,$0.961$] & 71 \\
Naive RAG (BM25-lite) & $0.500$ [$0.406$,$0.594$] & $0.221$ [$0.152$,$0.310$] & $0.868$ [$0.767$,$0.929$] & 68 \\
Oracle (ceiling) & $1.000$ [$0.964$,$1.000$] & $1.000$ [$0.964$,$1.000$] & $1.000$ [$0.947$,$1.000$] & 69 \\
\bottomrule\end{tabular}}
\end{table}

\begin{figure}[ht]\centering
\includegraphics[width=\textwidth]{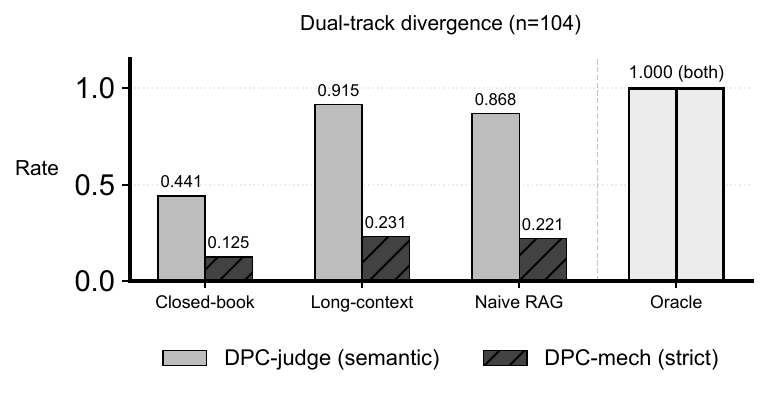}
\caption{The dual-track divergence. Semantic correctness (DPC-judge, light bars) masks mechanical-verification failure: among answers with valid judge verdicts, the text-augmented baselines achieve DPC-judge rates of $0.87$--$0.92$, while reproducing the golden fragments or evidence spans exactly on only $0.13$--$0.23$ of questions (DPC-mech, dark bars). The oracle ceiling scores 1.0 on both tracks, sanity-checking the metric logic.}\label{fig:dual_track_divergence}
\end{figure}

The long-context baseline scores a perfect judge rate on the time-constrained and multi-hop groups, so the released set leaves no headroom for the temporal-slice advantage H2 predicts; per-cell valid-$n$ denominators ship with the release. The long-context baseline's required-chapter range derives from the benchmark's gold annotation, so it receives the same gold-range information attributed to the chapter-scoped variant; both are labeled accordingly in the release, and neither is granted to the proposed system. Three measured facts stand out. First, parameter memory is strong and unverifiable: the closed-book LLM answers $44.1\%$ of questions correctly by the semantic judge, but its CF is strictly zero (no memorized answer is fully verbatim), and its $12$ MIXED failures show memorized fragments coinciding with the source; its $45$ no-verdicts share the judge's empty-output failure mode on the empty-evidence prompt, diagnosed for the oracle below. Second, the two text-augmented baselines overlap on both accuracy tracks (DPC-judge $0.915$ vs.\ $0.868$) and on CF ($0.510$ vs.\ $0.500$): no reliable difference is visible at this scale (McNemar on the $59$ questions with paired judge verdicts, $9$ discordant pairs, exact two-sided $p=0.18$); formal paired tests at campaign scale remain pre-registered---character-bigram retrieval brings no measurable citation gain. Third, the dual-track divergence is systematic (Figure~\ref{fig:dual_track_divergence}): long-context semantic correctness ($0.915$) exceeds strict fragment matching ($0.231$) by $68.4$ percentage points---strict fragment matching is a strict lower bound on answer correctness, not a citation measure. The few-shot demonstrations nevertheless matter: an exploratory zero-shot run on the 46-question preliminary set before the protocol freeze scored long-context CF $0.109$ ($5/46$) and naive-RAG CF $0.065$ ($3/46$), versus $0.510$ and $0.500$ with the frozen demonstrations---prompt-design context, not a second test-split evaluation.

\begin{table}[htb]
\caption{Long-context baseline by question group (descriptive; per-category cells are not hypothesis tests).}
\label{tab:secondary_results}
\centering\normalsize\setlength{\tabcolsep}{3pt}

\adjustbox{max width=\columnwidth}{%
\begin{tabular}{lcccc}\toprule
Group & $n$ & CF & DPC-mech & DPC-judge \\ \midrule
General & 81 & $0.481$ & $0.259$ & $0.902$ \\
Time-Constrained & 15 & $0.667$ & $0.133$ & $1.000$ \\
Space-Constrained & 8 & $0.500$ & $0.125$ & $0.875$ \\
\midrule
Single-hop & 88 & $0.511$ & $0.205$ & $0.900$ \\
Multi-hop & 16 & $0.500$ & $0.375$ & $1.000$ \\
\bottomrule\end{tabular}}
\end{table}

\begin{table}[htb]
\caption{Citation-faithfulness failure taxonomy (104 answers per baseline). OK: fully verbatim; MIXED: at least one non-verbatim fragment (paraphrase or prefix contamination); NON-VERBATIM: no verbatim fragment; ERROR: malformed output.}
\label{tab:cf_taxonomy}
\centering\normalsize\setlength{\tabcolsep}{3pt}

\adjustbox{max width=\columnwidth}{%
\begin{tabular}{lcccc}\toprule
Method & OK & MIXED & NON-VERBATIM & ERROR \\ \midrule
Closed-book LLM & 0 & 12 & 67 & 25 \\
Long-context LLM & 53 & 30 & 15 & 6 \\
Naive RAG & 52 & 28 & 15 & 9 \\
Oracle (ceiling) & 104 & 0 & 0 & 0 \\
\bottomrule\end{tabular}}
\end{table}

\begin{figure}[ht]\centering
\includegraphics[width=\textwidth]{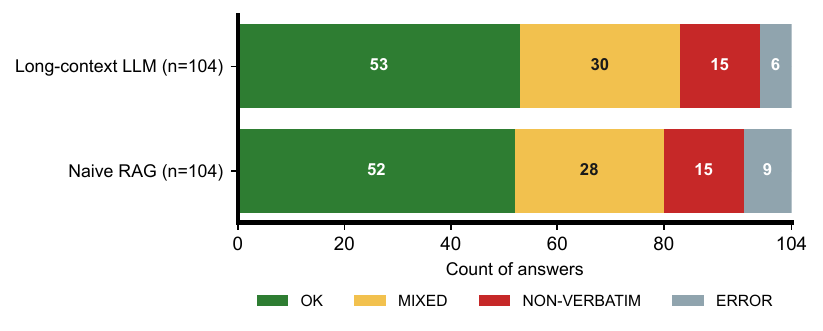}
\caption{Citation-faithfulness failure taxonomy for the text-augmented baselines. MIXED (partially verbatim, partially paraphrased evidence) is the dominant failure class, showing that LLMs default to rewriting even under few-shot verbatim enforcement; NON-VERBATIM and ERROR complete the loss.}\label{fig:cf_failure_taxonomy}
\end{figure}

The taxonomy of Figure~\ref{fig:cf_failure_taxonomy} shows where citation faithfulness is lost. The dominant failure class of both text-augmented baselines is MIXED: partially verbatim, partially rewritten evidence even under frozen few-shot enforcement---the behavior the per-fragment CF rule is designed to penalize; the closed-book model falls into NON-VERBATIM and ERROR, with no source to anchor its citations. Section~5 analyzes the mechanism behind the MIXED class and the dual-track divergence.

\begin{figure}[ht]\centering
\includegraphics[width=\textwidth]{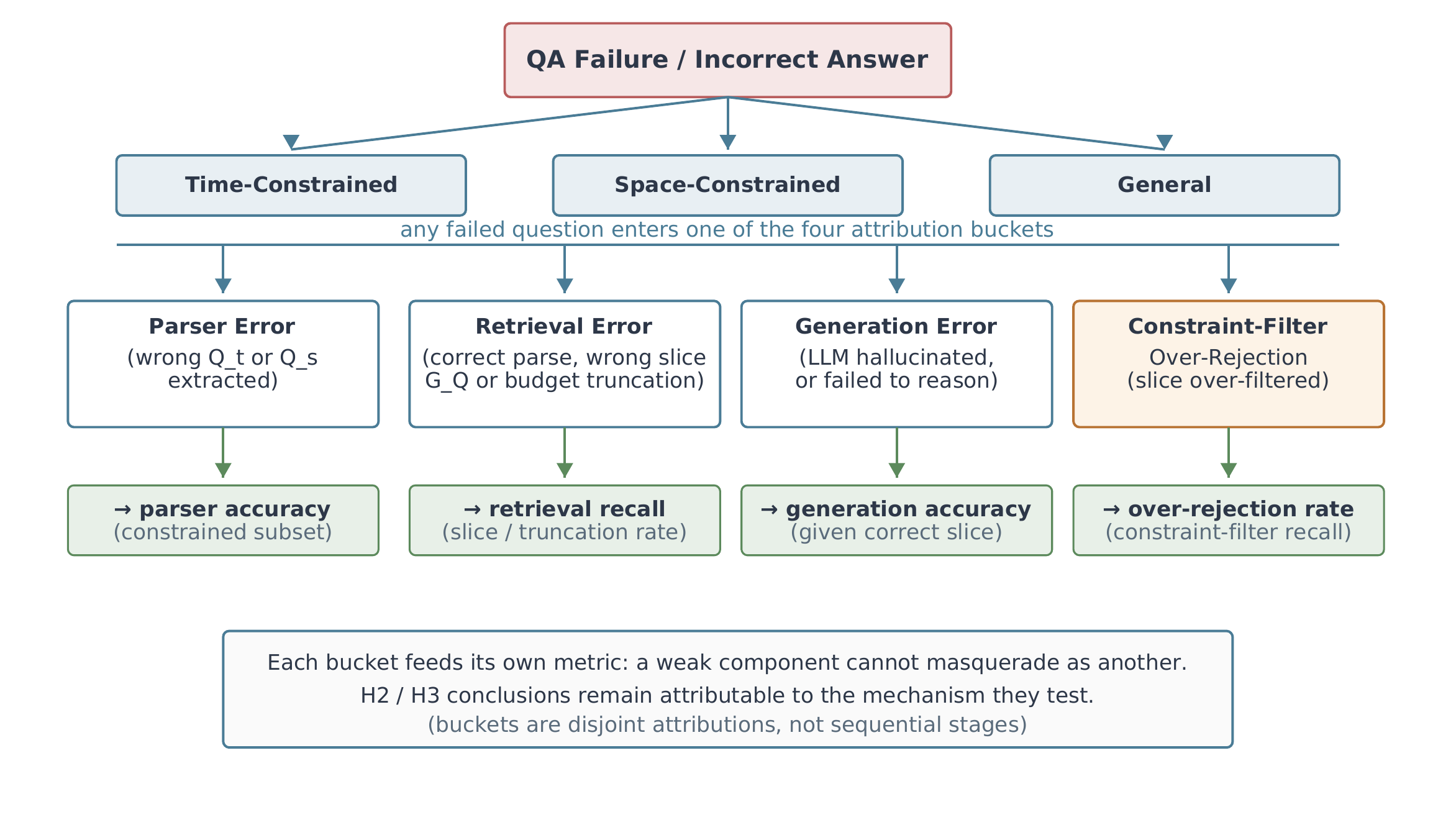}
\caption{Diagnostic protocol for error attribution. The root node is a QA failure. Level 1 stratifies the failed question by category: time-constrained, space-constrained, or general. Level 2 attributes the failure to one of four buckets: a parser error (wrong constraint $Q_t$ or $Q_s$ extracted), a retrieval error (correct parse, wrong slice $G_Q$, or budget truncation of an answer path), a generation error (correct slice, LLM hallucinated or failed to reason), or a constraint-filter over-rejection (a correct parse whose slice filter discards the answer path). Each bucket feeds its own metric, so a weak component cannot masquerade as another, and the H2/H3 conclusions remain attributable to the mechanism they test.}\label{fig:stratification_protocol}
\end{figure}
\subsection{Mechanism Pre-validation (Development Set)}
Before the full system is built, the retrieval-and-assembly mechanism it will inherit---query parsing, slice retrieval, and grounded generation in miniature---was pre-validated on the released development set under the diagnostic attribution protocol of Figure~\ref{fig:stratification_protocol}; the campaign's test split remained disjoint and unused. The miniature mechanism is deliberately zero-leak: it receives only the question text and the premise anchor, retrieves chapter blocks through premise- and question-anchored character-bigram windows, and assembles the answer by verbatim stitching under a two-step prompt plus a self-check correction round and a deterministic repair. Questions whose two question marks admit a clean split become two single-information-point sub-questions, merged. The pre-validation surfaced one benchmark-design finding. The golden answer parts are expert-phrased paraphrases: only $10$ of $104$ two-part golden answers can be reproduced verbatim from the source, so a strictly verbatim system faces a structural ceiling of $0.10$ on the original DPC-mech rule. The metric was amended to the union rule defined above; the amendment predates all baseline scoring reported in this paper, and every number herein is measured under the amended rule.
\begin{table}[htb]
\caption{Mechanism pre-validation on the released development set (104 questions). The first three rows are the development trajectory (steps bundle component changes); the final two rows are a controlled ablation isolating the deterministic repair. Development-set diagnostics, not campaign results.}
\label{tab:mechanism_ablation}
\centering\normalsize\setlength{\tabcolsep}{3pt}

\adjustbox{max width=\columnwidth}{%
\begin{tabular}{lccc}\toprule
Mechanism configuration & CF & DPC-mech & DPC-judge ($n$) \\\\ \midrule
Premise window only & $0.240$ & $0.144$ & --- \\\\
$+$ question-anchored windows & $0.538$ & $0.308$ & $0.875$ (64) \\\\
$+$ gate fix and split-question assembly & $0.779$ & $0.212$ & $0.892$ (65) \\\\
$+$ wider windows, no repair & $0.856$ & $0.317$ & $0.915$ (71) \\\\
$+$ deterministic repair (final) & $0.885$ & $0.413$ & $0.919$ (74) \\\\
\bottomrule\end{tabular}}
\end{table}
Three measured facts from the trajectory matter for the full campaign. First, question-anchored retrieval is the largest component: it raises the share of questions whose both evidence spans fall in the window from $11$ to $49$ of $104$, and CF from $0.240$ to $0.538$. Second, the controlled final pair isolates the deterministic repair: it adds $0.029$ CF and $0.096$ DPC-mech---a strengthening, not a decisive component; the mechanism confirms on all three gates even without it. Third, the residual failure class of the final configuration is paraphrase rather than missing evidence: $11$ of $104$ evidence outputs remain partially rewritten (MIXED) even under the frozen verbatim demonstrations---exactly the behavior the system's constraint layer is designed to attack at extraction time rather than answer time.

\subsection{System Construction Installment (Development Set)}

This installment reports the first full-corpus implementation of the currently realized configuration of the pipeline of Section~3---the modules exist as working code, not as design prose---together with development-set diagnostics; the design-to-implementation deltas are disclosed explicitly below. The campaign's test split remained disjoint and unused throughout the construction work, and the quantities the design assigns to expert verification are reported at the end of this subsection. The realization policy is deterministic-first: wherever the design allows a choice between a learned and a symbolic component, the symbolic one is built first, and the learned alternative is deferred to the full build, so every reported construction number below is reproducible by re-running the shipped pipeline. The only exceptions are the two passes that the design already assigns to the LLM: schema-prompted extraction and the event-timeline pass, each bounded at one call per chapter ($240$ calls in total) in the base build, with the repair loop set to $R_{max}=0$---deterministic repair only---for this build. The trajectory's density-50 build (R2 below) uses two-pass half-chapter extraction: each chapter body is split at a paragraph boundary near its midpoint and each half receives its own bounded call with a $25$--$40$ triple target, the halves merged with entity-id deduplication---two calls per chapter, a discrepancy against the base build's bound that is disclosed here rather than folded into the bound's wording.

Module A over $120/120$ chapters yields $3{,}347$ entity records in the base (density-20) build; the constrained files hold $2{,}497$ triples ($2{,}433$ located verbatim plus $64$ non-verbatim candidates). The constraint layer's construction-time value showed up immediately in an unexpected place. The extraction schema names the relation types and requests the evidence span with its chapter, but the prompt does not state which chapter is being extracted, and the schema example carries the literal field \texttt{chapter: 1}; the model accordingly copied the example value in $54$ of $120$ chapters, mislabeling $1{,}099$ triples (all of them to Chapter 1) while their evidence spans were otherwise verbatim. A deterministic repair pass---locate each evidence span verbatim in the source chapters and rewrite the coordinate from its true location---corrected all $1{,}099$ labels with zero LLM repair rounds, and the same pass doubles as the first half of the hallucination check: $64$ spans ($2.6\%$) match no chapter text verbatim and form the candidate list for the expert annotation slot below. The residual rule violations after the deterministic pass are $142$ ($1.2$ per chapter): $139$ type-signature mismatches, $2$ genealogy cycles, and $1$ master-servant conflict---the rule-candidate inventory that seeds the ontology acquisition protocol, whose expert verification (rule count and person-hours) is reported at the end of this subsection. These violating triples remain in the graph (the design's uncertainty-log exclusion for hard violations is not yet wired, and the soft low-confidence attribute marking is likewise deferred to the full build); the dense build's type-signature inventory is inflated by the resolved-entity-id convention, whose identifiers are canonical labels rather than type-prefixed---the inventory is recomputed under the resolved convention in the release.

Entity resolution in Module B is the realization's clearest design delta. The design assigns alias unification to LLM cluster proposals validated by two signals; the built version is a deterministic union-find over two overlap signals---shared canonical labels, and alias overlap with another entity's canonical label---which merged $1{,}105$ raw entity identifiers into $774$ unified entities ($-30\%$); the repaired triple set entering resolution held $2{,}497$ triples ($2{,}433$ located verbatim plus $64$ non-verbatim candidates), and resolution deduplicated it to $2{,}492$. The dev-corpus immediately produced a measured failure that motivated a guard rule: generic kinship and address terms (``aunt'', ``second master'', ``mistress'') used as aliases drove false merges across distinct servants and relatives (one cluster fused four minor characters onto a fifth; another fused two court ladies), so such terms are excluded from the merge signal while remaining available to retrieval. Three seed rename pairs that the source text states explicitly (the same character renamed mid-story, e.g., Mingyan/Baiming) are merged by fiat. The LLM disambiguation of residual ambiguous clusters remains deferred to the full build, and the merge-error measurement against the annotation subset is deferred to the full build.

The event-timeline pass extracted $1{,}989$ event occurrences across the corpus, with the chapter number injected into the prompt---the root-cause fix for the copied-coordinate failure above---and assembled them deterministically: participants resolve to unified entity identifiers through the alias index with canonical-name priority and chapter-co-occurrence tie-breaking, which reduced unresolved participant mentions from $156$ ($3.4\%$) to $58$ ($1.3\%$ of $4{,}577$ mentions; the remainder are collective nouns and minor characters absent from extraction), and $4.9\%$ of event evidence spans are non-verbatim candidates for the annotation slot. The pass yields $14$ anchor-event candidates, each with its chapter span; the domain-expert confirmation of the anchor list is reported here rather than left pending: $8$ candidates were confirmed and $2$ were refined (the decree ordering the Imperial Consort's visit was downgraded in favor of the visit itself, Chapters 17--18; the proposal to raid Prospect Garden was downgraded in favor of the raid itself, Chapter 74), yielding a confirmed anchor set of twelve events spanning Chapters 13 to 120, after the three deferred candidates were resolved by the same expert: Baoyu's beating (Chapter 33) confirmed, Granny Liu's second visit (Chapter 39) downgraded (literary importance without temporal-reference function), and Qingwen's death (Chapter 77) confirmed on the verbatim in-text reference ``the year Qingwen died'' used by a later chapter to locate past time---the operational definition of an anchor made explicit by the evidence itself. Narrative time is realized as the pair (chapter, sequence position) plus anchor references; the designed Allen-interval refinement, the snapshot-plus-delta storage model, and the temporal-consistency rule family of Section~3 remain deferred to the full build, and the coarse chapter index already supports the slice lookup the queries below exercise.

Module C is realized as a pure deterministic retriever over the versioned slices: evidence spans are scored by the number of question entities they mention (the span-level both-in signal), edges are grouped by chapter and assembled in descending chapter score (narrative locality), free question keywords outside the entity vocabulary contribute additional span hits, and a timeline lookup maps a question's anchor-event mention, together with its temporal connective, to a chapter window ($\le$ the anchor chapter for ``before'', $\ge$ for ``after''). The designed graph-distance ranking with evidence-recency tie-breaking, the shortest-path retention, and the scene-to-entity spatial index of Section~3 are not yet built: the realized retriever substitutes span-neighborhood BM25 ranking (R3--R5 below), disclosed here as a design delta in the same style as the union-find substitution above. Table~\ref{tab:system_dev} reports development-set diagnostics on six questions under the realized configurations.

\begin{table}[htb]
\caption{Development-set diagnostics on the constructed system. Retrieval width is the number of supporting triples selected under the token budget; CF is citation faithfulness. Six development questions (the three smoke questions of the mechanism pre-validation above and three time-constrained questions whose answers were source-verified before the runs, selected as the only dev questions with timeline-window triggers at selection time); configurations: v1 (entity-endpoint retrieval), v2 (the realized DSGR retriever), and timeline on/off for the time-constrained questions. Development diagnostics, not campaign results.}
\label{tab:system_dev}
\centering\normalsize\setlength{\tabcolsep}{3pt}

\adjustbox{max width=\columnwidth}{%
\begin{tabular}{llccc}\toprule
Question (type) & Configuration & Width & CF & Verdict \\ \midrule
RCQ-P003 (two-part) & v1 / v2 & $56$ / $51$ & $1.0$ / $1.0$ & correct \\
RCQ-P15 (two-part) & v1 / v2 & $39$ / $38$ & $1.0$ / $1.0$ & correct \\
RCQ-MH-005 (multi-hop) & v1 / v2 & $942$ / $71$ & $1.0$ / $1.0$ & correct \\
TQ-1 (time, ``after'') & timeline on / off & $11$ / $20$ & $0.0$ / $0.0$ & wrong (recall gap) \\
TQ-2 (time, ``before'') & timeline on / off & $29$ / $53$ & $1.0$ / $1.0$ & incomplete (recall gap) \\
TQ-3 (time, ``after'') & timeline on / off & $78$ / $74$ & $1.0$ / $1.0$ & correct / wrong \\
\bottomrule\end{tabular}}
\end{table}

Three measured facts from the diagnostics matter for the full campaign. First, the realized retriever collapses the retrieval width of the multi-hop smoke question from $942$ to $71$ supporting triples ($-92\%$) without losing the answer, which is the quantity the full campaign's retrieval-efficiency comparison measures at scale. Second, \begin{figure}[ht]\centering
\includegraphics[width=\textwidth]{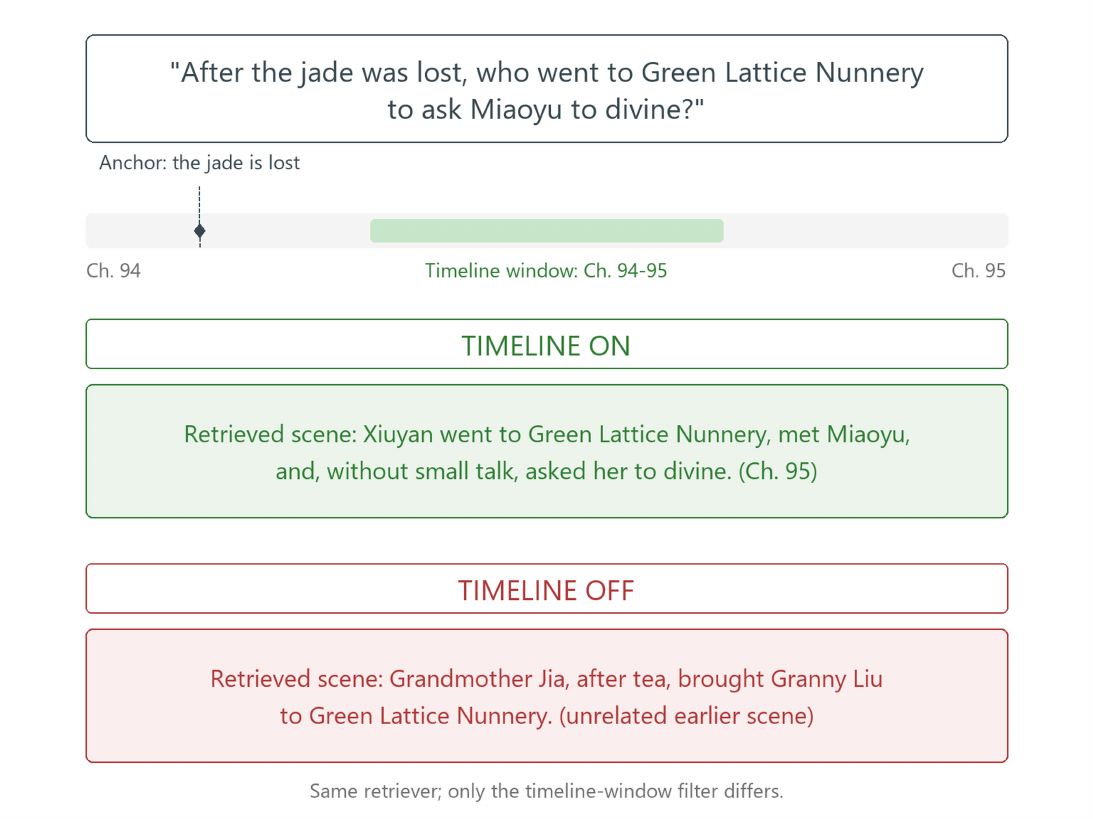}
\caption{The timeline-window diagnostic on TQ-3: the same retriever, with the timeline window on, returns the gold evidence neighborhood (Xiuyan asking Miaoyu to divine, Chapter 95); with the window off, it returns an unrelated earlier scene (the Grandmother Jia and Granny Liu visiting Green Lattice Nunnery). Same retriever, only the timeline window filter differs.}\label{fig:tq3_ablation}
\end{figure}

the dual-coordinate mechanism has a single-question development-set diagnostic (not an ablation at scale): on TQ-3 (``after the jade was lost, who asked Miaoyu to divine?'') the timeline window plus free-keyword scoring returns the gold evidence span, while the same retriever without the timeline window answers from an unrelated earlier scene (Figure~\ref{fig:tq3_ablation})---the exact behavior the temporal slice is designed to prevent, observed on the built system rather than asserted from the design. Third, the recall ceiling is visible at development scale: on TQ-1 and TQ-2 the gold evidence spans are absent from the extracted graph entirely, because chapter-level extraction emits roughly twenty triples per chapter and both answers fall outside what that cap captured. This bounds the system's answerable set by extraction completeness rather than retrieval or generation quality, and it motivates the extraction-density ablation that the full campaign pre-registers: the same pipeline under a raised per-chapter extraction budget, isolating recall from the routing mechanism the hypotheses test.

\subsection{System Iteration Trajectory (Development Set)}

\begin{figure}[ht]\centering
\includegraphics[width=\textwidth]{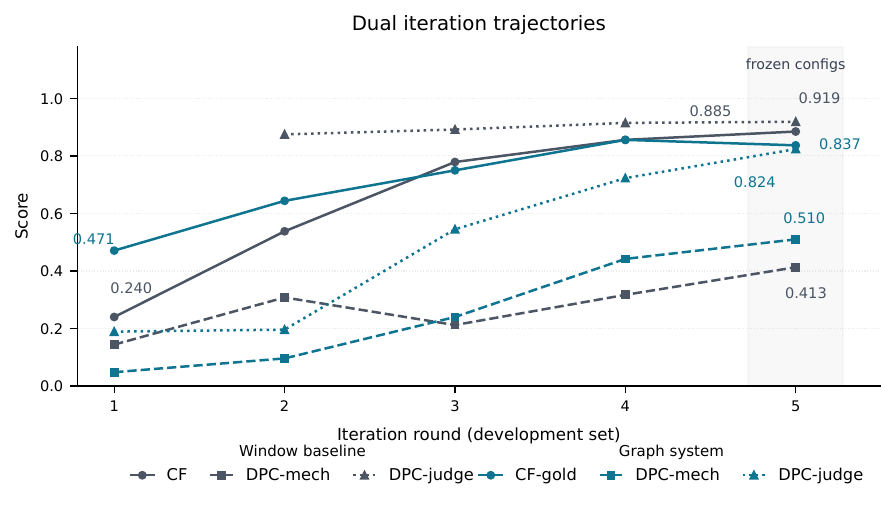}
\caption{Dual iteration trajectories on the same development set: the window baseline's five mechanism configurations (Table~\ref{tab:mechanism_ablation}) and the system's five recorded retrieval rounds (Table~\ref{tab:system_trajectory}). The interventions differ by construction---window composition for the baseline, graph retrieval for the system; only the final configurations are frozen for the campaign, and the system's frozen R5 context is about seven thousand tokens of graph-anchored neighborhoods against the baseline's two-thousand-token window.}\label{fig:trajectory_dual}
\end{figure}

The iteration budget was set at four rounds before iteration began (the release's commit history timestamps the commitment); R5 is the disclosed extension beyond that budget, reported in the same table rather than folded into a round. The extension was authorized after R4's measured judge gap ($0.723$ vs.\ $0.919$) made the ranking-side residual visible; the decision and its timestamp are recorded in the release, and the frozen R5 configuration is flagged as the development-set selection on the DPC-mech track (CF-gold is non-monotone: R4's $0.856$ exceeds R5's $0.837$), not a test-set estimate. Every round below is a recorded configuration change over the frozen assembly machinery---the development-set counterpart of the mechanism trajectory of Table~\ref{tab:mechanism_ablation}, whose five rows the window baseline consumed on the same set. The two trajectories are presented together (Figure~\ref{fig:trajectory_dual}) so that neither system's development history is hidden; the interventions differ by construction (window composition for the baseline, graph retrieval for the system) with one deliberate homology: the R5 union of quote-stripped sub-question queries applies at retrieval time the same two-question-mark decomposition the frozen assembly already applies at generation time, so the mechanism is format-agnostic rather than a per-question patch, and its isolated contribution is reported in the release's component ablation. Only the final configurations are frozen for the campaign.

\begin{table}[htb]
\caption{System iteration trajectory on the development set (104 questions). Each round changes one retrieval mechanism over the frozen assembly (component bundles noted per round); the MVP v6 row is the window baseline's frozen configuration from Table~\ref{tab:mechanism_ablation}. The window baseline's mechanism receives the benchmark's verbatim premise anchor, while the system's retrieval receives only the question text; the asymmetry is disclosed rather than hidden, and the query-side premise-injection negative result below does not settle the window-side question---the campaign instead removes the premise anchor from both arms, a configuration delta disclosed in the Main Campaign subsection, and the chapter-scoped condition remains scheduled with the baseline family. CF-gold is citation faithfulness scored against the gold chapter range; DPC-judge is computed over its own valid-$n$, with no-verdict-as-incorrect worst-case bounds of $0.173$/$0.192$/$0.529$/$0.702$/$0.808$ for R1--R5 (the window baseline's is $0.654$ over its $74$ valid verdicts; its gold-range CF is $0.894$ and its context-track CF $0.885$). Development diagnostics, not campaign results.}
\label{tab:system_trajectory}
\centering\normalsize\setlength{\tabcolsep}{3pt}

\adjustbox{max width=\columnwidth}{%
\begin{tabular}{llccc}\toprule
Round (intervention) & CF-gold & DPC-mech & DPC-judge & valid $n$ \\ \midrule
R1: entity-endpoint retrieval, density 20 & $0.471$ & $0.048$ & $0.189$ & 95 \\
R2: density 50 (two-pass half-chapter) & $0.644$ & $0.096$ & $0.196$ & 102 \\
R3: span-neighborhood BM25 ranking & $0.750$ & $0.240$ & $0.545$ & 101 \\
R4: select-short expand-wide (BM25 over $\pm$150 selection units, $\pm$400 context units), adjacent-chapter pool, Chinese-numeral parser fix (chapter 116 was parsed as 1016) & $0.856$ & $0.442$ & $0.723$ & 101 \\
R5: union of whole-question and quote-stripped sub-question BM25 queries, context units $\pm$400 to $\pm$300 & $0.837$ & $0.510$ & $0.824$ & 102 \\
\midrule
MVP v6 (window baseline, frozen) & --- & $0.413$ & $0.919$ & 74 \\
\bottomrule\end{tabular}}
\end{table}

Three findings from the trajectory shape the campaign. First, the extraction-density rounds and the ranking rounds decompose cleanly: R2 (density $20 \to 50$) moves CF-gold by $+0.17$ while DPC-mech moves by only $+0.05$, because the retrieved evidence lands in the right chapters but the ranking still selects the wrong spans; R3 (neighborhood BM25) moves DPC-judge by $+0.36$ at one stroke, because the ranking problem was the binding constraint all along. Second, the benchmark's shortcut-free construction is a hard lexical wall at span granularity: the expert rewrite removes surface overlap between question and answer passage, so a BM25 ranker over the graph's verbatim evidence spans puts all golden spans in context for $0.067$ of questions, while the same ranker over graph-anchored neighborhoods of the spans reaches $0.596$---the passage around a relation carries the lexical signal the relation itself was stripped of. Third, at the frozen R5 configuration the system reaches CF $0.913$ on the context track against the baseline's $0.885$ ($+2.8$ points) while its gold-range track $0.837$ trails the baseline's gold-range $0.894$ by $5.7$ points, and DPC-mech $0.510$ exceeds the baseline's $0.413$ by $9.7$ points, with a context of about twenty graph-anchored neighborhoods expanded to roughly twelve thousand characters (about seven thousand tokens) against the baseline's two-thousand-token contiguous window---larger in raw size, but every selected unit is graph-anchored, provenance-complete, and temporally scoped, the properties the provenance and temporal-coordinate metrics of the design measure directly, while DPC-judge remains $0.095$ below the baseline ($0.824$ vs.\ $0.919$; the no-verdict-as-incorrect bounds invert, $0.808$ vs.\ $0.654$, because the system's valid-$n$ is $102$ against $74$). The gap is measured, not asserted: R5's per-category judge rates are General $0.861$ ($n{=}79$), Space-Constrained $1.000$ ($n{=}8$), and Time-Constrained $0.533$ ($n{=}15$), so the deficit concentrates in the time-constrained category, where the gold spans split across adjacent chapters and the second span is buried deep in ranking. The time-constrained evolution question P15 is a concrete example. Although its diagnostics-table verdict is ``correct'' under v1/v2, this result does not transfer to R5, where its two gold spans rank 59th and 121st in the retrieval pool. A residual set of questions shows the same coverage problem because their golden passages sit in scenes the extraction covers only sparsely. Negative results recorded along the trajectory, printed for the same reason as the table (premise-anchoring failure added, measured at $0.167$ on questions with non-empty premises): premise injection into the query dilutes the BM25 signal ($-0.03$), stripping interrogatives from the query hurts ($-0.03$), enlarging neighborhoods at fixed budget hurts ($-0.12$), and premise-anchored neighborhood selection---locating the premise in the source and preferring nearby edges---recovers only $0.167$ of questions because the answers of multi-chapter questions live outside the premise's chapter; each measured on the development set and each reproduced by the shipped scripts. Together the negatives teach one lesson: the evidence anchor is the correct selection unit, the query's non-entity text the correct ranking signal, and coverage must come from the passage around the anchor rather than the query's own position.

The expert-verification results are reported here rather than left pending, in protocol order. The anchor-event confirmation is reported above (twelve anchors). The ontology acquisition protocol over the $142$-violation inventory yielded seven verified rules---the three seed families plus alias-ownership disambiguation, cross-type merge prohibition, verbatim multi-span evidence integrity, and an anonymous-collective entity threshold---at an estimated expert labor of eight hours; the same verification flagged five erroneous merge clusters (including one that fused three distinct princely titles) and twenty-one alias-ownership conflicts, each with a disposition. The golden annotation of Chapters 3, 8, and 20 is likewise reported rather than left pending: over 98 annotated items (44 triples and 54 events), 71 pass, 10 violate the annotation (strict hallucination rate 10.2\%, 95\% Wilson interval [5.0\%, 17.8\%]), and 17 need presentation-level corrections, for an any-correction rate of 27.6\%. The decomposition matters more than the single figure: relation-level errors are 5.1\% (5/98, including one case where the extraction inverted two characters' sleeping quarters against the source), evidence-integrity errors 4.1\% (4/98 non-verbatim spans), and type-signature errors 1.0\% (1/98)---the non-verbatim class being the measured justification for the design's evidence-anchoring soft constraint, and the any-correction rate being read primarily as a predicate-precision indicator (assigned-to overuse on person--person pairs, over-specific scene naming) rather than a maintenance cost, though it is both. Two caveats attach to the rate: the three annotated chapters are kinship- and social-narrative dense and do not represent poetry, banquet, or descriptive chapter types, so the rate is a measured point on an unrepresentative sample; and the rate is the post-deterministic-repair condition with the LLM repair loop at $R_{max}=0$. The H1 comparison is now measured at extraction level on the same subset against the unconstrained (raw) extraction, whose files ship with the release: all $44$ raw triples carry wrong chapter coordinates (the copied-coordinate class, $44/44$), which the constraint layer's deterministic pass repairs completely, and $6$ of $98$ raw items carry non-verbatim evidence against $4$ post-repair (the four residuals are the gold's own evidence-integrity errors, whose spans match no chapter text and are therefore unrepairable by location). The ontology rules, however, flag none of the ten gold-violating relations ($0/10$): the gold's error classes---in-law as spouse, grandparent as parent, sibling for niece, visitation to a person, residence inversion, servant as fellow-servant's master---fall outside the current rule families. H1's extraction-level verdict is therefore split: the layer eliminates the coordinate class entirely, while on the rejection side the current rule families would not have flagged any of the ten gold violations, so the rejection component's value on the gold's error classes is undemonstrated rather than refuted---the measured basis for both the auditability reading of the layer and the per-relation-type rule-coverage tracking countermeasure; the active-repair condition remains the scheduled full-pipeline measurement. Entity-merge error measurement is outside the present measured installment and is therefore excluded from the current claims. These measurements are reported only within the scope of the present construction installment and do not extend to the deferred active-repair or entity-merge evaluations.

\subsection{Main Campaign (Held-out Test Split)}

The campaign's held-out test split is reported here: 120 questions over the 60 chapters disjoint from the preliminary release's 60, used exactly once under the frozen R5 configuration; the split ships with the release (question IDs, chapters, the adjudicated label file, raw answers), and the leakage registry marks these 60 chapters as the test split. The split was built under the same construction protocol as the preliminary set---LLM-assisted candidate generation from chapter evidence, expert rewriting of every shortcut-admitting question, the four-check construction rule set at $0/120$ violations, and verbatim evidence spans of at least ten characters---so the $0\%$ shortcut rate holds by construction on both sets; the adjudication pass is specific to this split, and the preliminary release's labels predate it. A second annotator adjudicated the expert's difficulty labels under two stated criteria: multi-hop requires a reference (identity, kinship, or addressee) that the question text does not name, and a hard question is one whose two evidence spans sit on opposite sides of a line break (the digitization's single-newline block boundary, which the rule set uses as its paragraph proxy) at least ten sentence boundaries apart, or whose unnamed-reference resolution has at least two targets. Both criteria are mechanical---span geometry and question-text reference checks---and were applied blind to every measured score; the label file ships with the release, recording the expert's demotion rationale per question. No question was removed: the split holds all 120. The adjudicated split holds 111 medium and 9 hard questions, 116 single-hop and 4 multi-hop (one of the four multi-hop questions is also hard), with category counts General 42, Space-Constrained 33, Time-Constrained 45. The split is medium-heavy by construction: the adjudication demotes exactly the questions whose second answer was reachable from the first, so the split measures direct-fact retrieval plus the nine hard and four multi-hop questions that survived. Both retrieval arms receive only the question text---removing the development-set premise-anchor asymmetry; the baseline's window now derives from the question text alone, a condition never measured on the development set and disclosed as a configuration delta rather than folded into the frozen label. The closed-book arm receives the identical prompt and frozen demonstrations with an empty context; temperature, judge model, and retry protocol match the preliminary release.

\begin{table}[htb]
\caption{Main campaign on the held-out test split (120 questions), with 95\% Wilson intervals. CF is the gold-range track (the development-set CF-gold column). DPC-judge is computed over answers with a verdict after three retries (valid $n$); counting no-verdicts as incorrect gives worst-case judge bounds $0.258$/$0.850$/$0.858$ for the three arms (the system's single no-verdict, TS76-04, a nested-quote multi-hop question, has an empty answer from a transient rate-limit error (429); the closed-book arm's seven no-verdicts are retry exhaustions with empty verdict content on empty-evidence questions, the failure mode diagnosed on the preliminary set; transcripts ship with the release). For the closed-book arm the CF check runs over the answer text itself---no evidence field exists---so its three passes are fully verbatim answers reproduced from memory, not citations.}
\label{tab:campaign_main}
\centering\normalsize\setlength{\tabcolsep}{3pt}

\adjustbox{max width=\columnwidth}{%
\begin{tabular}{lcccc}\toprule
Method & CF & DPC-mech & DPC-judge & \begin{tabular}{@{}c@{}}DPC-judge\\ valid $n$\end{tabular} \\ \midrule
Closed-book LLM & $0.025$ [$0.009$,$0.071$] & $0.083$ [$0.046$,$0.147$] & $0.274$ [$0.201$,$0.363$] & 113 \\
MVP v6 (window baseline, frozen) & $0.817$ [$0.738$,$0.876$] & $0.675$ [$0.587$,$0.752$] & $0.850$ [$0.775$,$0.903$] & 120 \\
NS-ST-GraphRAG v7 (frozen R5) & $0.825$ [$0.747$,$0.883$] & $0.733$ [$0.648$,$0.804$] & $0.866$ [$0.793$,$0.916$] & 119 \\
\bottomrule\end{tabular}}
\end{table}

Three measured facts shape the campaign's reading. First, the retrieval stack accounts for the entire advantage over parametric memory: the closed-book arm answers $27.4\%$ of questions fully correctly by the judge and $8.3\%$ by mechanical reproduction, so the system's advantage over the memory floor is $+0.58$ judge on the $112$ paired verdicts ($96$ against $31$) and $+0.650$ mech; the answering pipeline's residual ceiling is not partitioned here, since no held-out oracle arm ran. The closed-book leakage is unevenly distributed: judge $0.41$ [$0.271$,$0.566$] on General ($n{=}39$) against $0.226$ [$0.114$,$0.398$] on Space-Constrained ($n{=}31$) and $0.186$ [$0.097$,$0.326$] on Time-Constrained ($n{=}43$), while closed-book mechanical reproduction on Time-Constrained is strictly zero ($0/45$)---the split's time questions are mechanically unreproducible from memory, which is the property the temporal-slice mechanism exists to exploit. Second, the system matches the window baseline with a directionally favorable but non-significant mechanical edge: mech $0.733$ vs.\ $0.675$ (McNemar, $10$ v7-only-correct vs.\ $3$ v6-only-correct discordant pairs, exact two-sided $p{=}0.092$; uncorrected chi-square $p{=}0.052$, continuity-corrected $0.096$), judge $0.866$ vs.\ $0.850$ ($7$ vs.\ $6$ discordant over the $119$ paired verdicts, exact $p{=}1.0$; the system's single no-verdict is excluded and counted against it in the worst-case bound), and CF-gold $0.825$ vs.\ $0.817$. The baseline's per-category mech rates are General $0.643$ ($27/42$), Space-Constrained $0.727$ ($24/33$), Time-Constrained $0.667$ ($30/45$), so the deltas are General $+0.071$, Space-Constrained $0.000$, Time-Constrained $+0.089$---the point estimates favor the graph arm when no scene anchor is present and are tied when one is present; this is a descriptive pattern, not a measured attribution. The pattern bears directly on the pre-registered hypothesis H2: by its falsification condition the constrained-category differential gain must exceed the general gain, and the delivered comparison does not meet it---the pooled constrained delta is $+0.051$ ($78$ questions) against $+0.071$ on General, with Space-Constrained zero. Time-Constrained alone is positive, but H2's condition is stated over the constrained pair, so the delivered campaign does not support H2, the chapter-scoped Naive RAG variant that completes its second prong remains scheduled, and no mechanistic claim is made from these numbers. Third, the development-set judge deficit of R5 does not transfer, and no cross-set delta is computed: the splits differ in composition by construction (63 hard against 9), and the window baseline's development-set runs held the premise anchor both campaign arms now forgo---a confound that would make any cross-set gap uninterpretable. Within the held-out split alone, the frozen configuration shows no reliable difference from the window baseline in either direction on the judge track and a directionally favorable, non-significant mechanical point estimate---this supports only a stability interpretation.

\begin{table}[htb]
\caption{Held-out campaign by question group (descriptive; per-group cells are not hypothesis tests). The difficulty adjudication shows directional separation at modest power: hard and multi-hop questions cut mechanical reproduction by roughly $0.3$--$0.5$ directionally (Fisher exact $p{=}0.056$ and $p{=}0.058$), each just outside the conventional threshold at the split's small cell sizes.}
\label{tab:campaign_stratification}
\centering\normalsize\setlength{\tabcolsep}{3pt}

\adjustbox{max width=\columnwidth}{%
\begin{tabular}{lccccc}\toprule
Group & $n$ & CF & DPC-mech & DPC-judge & valid $n$ \\ \midrule
General & 42 & $0.690$ [$0.540$,$0.809$] & $0.714$ [$0.564$,$0.828$] & $0.805$ [$0.660$,$0.898$] & 41 \\
Space-Constrained & 33 & $0.848$ [$0.691$,$0.933$] & $0.727$ [$0.558$,$0.849$] & $0.848$ [$0.691$,$0.933$] & 33 \\
Time-Constrained & 45 & $0.933$ [$0.821$,$0.977$] & $0.756$ [$0.613$,$0.858$] & $0.933$ [$0.821$,$0.977$] & 45 \\
\midrule
Hard & 9 & $0.667$ [$0.354$,$0.879$] & $0.444$ [$0.189$,$0.733$] & $0.625$ [$0.306$,$0.863$] & 8 \\
Medium & 111 & $0.838$ [$0.758$,$0.895$] & $0.757$ [$0.669$,$0.827$] & $0.883$ [$0.810$,$0.930$] & 111 \\
\midrule
Multi-hop & 4 & $0.250$ [$0.046$,$0.699$] & $0.250$ [$0.046$,$0.699$] & $0.333$ [$0.061$,$0.792$] & 3 \\
Single-hop & 116 & $0.845$ [$0.768$,$0.900$] & $0.750$ [$0.664$,$0.820$] & $0.879$ [$0.808$,$0.927$] & 116 \\
\bottomrule\end{tabular}}
\end{table}

The stratification is the most informative result of the held-out split. The four multi-hop questions cut the system to mech $0.250$ and judge $0.333$ ($n{=}3$) against $0.750$/$0.879$ on single-hop, and the nine hard questions to $0.444$/$0.625$ against $0.757$/$0.883$ on medium---the adjudicated labels separate performance directionally on both axes (Fisher exact $p{=}0.056$ and $p{=}0.058$), so the difficulty axis shows a consistent directional pattern at the split's small hard and multi-hop cells, and the multi-hop shortfall is the headroom the design's constraint-layer and temporal-slice ablations target, though none of H1--H4 takes multi-hop accuracy as its dependent variable, and follow-up must use development data under the once-only commitment. Two disclosures bound the comparison: the splits differ in composition by construction (63 hard on the preliminary release against 9 here), so cross-set totals are not compared---all campaign statements are same-set---and the system's single judge no-verdict (a nested-quote question) is counted as incorrect in the worst-case bounds. The human expert baseline pre-registered in the design---experts who did not annotate the gold set answering open-book with a time budget---remains a scheduled slot of the complete release, as do the four pre-registered baseline-family comparisons (Naive Vector RAG, GraphRAG, LightRAG, HippoRAG), the chapter-scoped Naive RAG variant (H2's second prong), the per-category McNemar comparisons with Holm-Bonferroni adjustment, the H1 active-repair and H3/H4 measurements, the extraction-quality, latency, and construction-cost metrics, and the full-corpus generalization measurement.

\section{Discussion and Analysis}
\subsection{Hypothesis-Wise Expectations}
The experimental design tests four hypotheses, stated here with their falsification conditions. H1: the neuro-symbolic constraint layer reduces the extraction-level hallucination rate relative to unconstrained extraction; the hypothesis is falsified if the full-campaign ablation shows no difference between the full system and the condition without the layer. H2: temporal-slice retrieval improves accuracy on constrained questions more than on general ones; the hypothesis is falsified if the full-campaign differential gain (proposed system minus baseline) on the time- and space-constrained categories is no larger than the differential gain on the general category, or if the chapter-scoped Naive RAG variant matches the proposed system on the constrained categories. H3: slice retrieval lowers retrieval latency relative to full-graph retrieval and remains competitive as chapter count grows, measured by construction cost and latency at progressive sub-corpus sizes, half, three-quarters, and the full corpus; the hypothesis is falsified if the latency advantage is absent at any measured scale, and the construction-cost component is falsified if slice-index construction exceeds full-graph construction cost at any measured scale. H4: the versioned graph reproduces the known narrative trajectory of the case study; the hypothesis is falsified if the evolution analytics contradict the power-decline baseline that literary scholarship describes. These interpretive commitments were specified before the held-out campaign; their measured status is now as follows: H1 is tested at the extraction level in Section~4 with a split verdict (the coordinate class eliminated, the rejection rules flagging none of the ten gold violations); H2's constrained-category condition is evaluated on the delivered campaign as unmet and disclosed; H3 and H4 remain untested under their falsification conditions. H1 concerns the constraint layer: if the full-campaign ablation shows that removing the layer raises the hallucination rate, the interpretation is that ontology rules catch a class of errors that no prompting discipline alone prevents; if the effect is small, the layer's value shifts to auditability rather than accuracy, and the construction-cost column decides the economic half of the question. H2 concerns temporal slicing: an accuracy gain concentrated in the time- and space-constrained categories, with no corresponding gain in the general category, is the pattern that attributes the improvement to the temporal machinery; a gain spread evenly across categories would instead suggest the advantage lies elsewhere, and the error-attribution protocol would then tell us where. H3 concerns efficiency: lower latency for slice retrieval relative to full-graph retrieval, with token overhead reported, establishes that temporal indexing is an efficiency mechanism rather than an additional cost center. H4 concerns evolution analytics and is treated in the case study below.

\subsection{Measured Failure Analysis (Preliminary Installment)}
The preliminary evaluation measures three failure mechanisms. The first is the MIXED citation class, the dominant failure mode of both text-augmented baselines (Table~\ref{tab:cf_taxonomy}): $30$ of $104$ long-context and $28$ of $104$ naive-RAG answers carried partially verbatim, partially rewritten evidence under frozen verbatim-extraction demonstrations. The mechanism is default-to-paraphrase behavior the prompt does not suppress; the per-fragment CF rule converts it into a measurable failure, and the few-shot sensitivity is measured directly (CF $0.109$ zero-shot vs.\ $0.510$). The second is parameter memory without grounding: the closed-book baseline answered $44.1\%$ correctly with no source text, and its $12$ MIXED failures show memorized fragments coincidentally matching the source---unverifiable ``evidence''. The third is the dual-track divergence: the long-context baseline answered $0.915$ semantically while reproducing the golden fragments or evidence spans exactly on only $0.231$. Free-text answers rarely reproduce golden fragments verbatim, so strict matching is a strict lower bound on correctness; the $0.915$--$0.231$ divergence quantifies the gap, and reporting both tracks keeps either from being read alone. The semantic track was validated on a $10$-question pilot (agreement $0.80$--$0.90$, Cohen's $\kappa$ $0.41$--$0.78$, zero gold misjudgments). Finally, the authors' own early free-generation pilot runs (archived with the release) produced shortcut rates of roughly $0.7$--$0.8$ under self-reported checks, while the expert-rewrite plus mechanical-assembly pipeline produced the released set's $0\%$ rate. That comparison is a pipeline replacement, not a component ablation: it does not isolate mechanical verification from human adjudication, but it shows the combined construction method reaches zero shortcuts.

\subsection{Measured Failure Analysis (Held-out Campaign)}
The held-out campaign's failures are attributed mechanically: for each of the 120 answers, the two golden evidence spans are checked as verbatim substrings of the answer's cited evidence (the deterministic provenance check the design pre-registers), and the evidence itself is checked against the gold chapter range, which separates the pipeline's stages without any model call. Four classes cover the 119 answers with classifiable cited evidence; the 120th (TS76-04, a nested-quote multi-hop question) carries an empty answer because its generation call failed with a transient rate-limit error (429), leaving no cited evidence to attribute and no judge verdict, and it is counted incorrect under the worst-case rule. Retrieval complete: both spans sit verbatim in the cited evidence (35 questions; 30 reproduce mechanically and the judge scores all 35 correct). Partial evidence: exactly one span is cited (37 questions), losing the second span $23$ times against the first's $14$---the second span ranks deeper than the first, transferring the development-set finding to the fresh split. Verbatim wrong passage: the cited evidence is fully verbatim inside the gold chapter range but contains neither golden span (31 questions; the answer is nevertheless mechanically correct on $26$ and semantically correct on $0.887$), a citation-misalignment class that the per-fragment CF rule cannot see, since it checks verbatimness rather than span identity---the exact blind spot the design's provenance tracing metric exists to close. Retrieval miss: the evidence contains neither span and is not fully verbatim in the gold range (16 questions, semantic accuracy $0.375$); the retriever's chapter scope always covers the gold chapters (zero scope misses across all 120), so the loss is ranking and budget truncation, not chapter routing. The mechanical losses ($32$ of $120$) decompose accordingly: $12$ retrieval misses, $9$ one-part paraphrases, $5$ assembly drops in which both spans sit in the cited evidence but the merged answer loses their verbatim form (all five judge-correct), $5$ wrong-passage citations whose answers also fail mechanically, and the single no-verdict---combining the nine one-part paraphrases with the five assembly drops, generation-side loss of verbatim answer form accounts for 14 of the 32 mechanical failures, slightly more than the 12 retrieval misses, and the difficulty axis concentrates at the retrieval stage: two of the four multi-hop questions (both medium) and two of the nine hard questions are retrieval misses, against twelve further medium questions.

\subsection{Anticipated Failure Modes of the Full Campaign}Three failure modes of the full campaign are committed to measurement in advance, and the measurement plan is part of the design rather than an afterthought of the results write-up. Ontology incompleteness: the rule set is verified over a development split, but a novel violation pattern absent from the rules passes unchecked, which bounds the claim that can be made about the hallucination rate and motivates the uncertainty log that preserves flagged-but-unverifiable triples for expert review. Two countermeasures address incompleteness itself rather than its measurement: rule coverage is tracked per relation type, so relation types with thin rule support are visible in the release, and a post-hoc expert spot-check samples accepted triples on the evaluation split to estimate the residual uncaught hallucination rate. The disjoint annotation discipline is a third, separate countermeasure: the hallucination metric is measured on data the rule authors never saw, so rule overfitting to the development split shows up as a metric gap rather than hiding. The generalization corpus serves as the out-of-distribution test of rule transfer: rules mined from the errors of a classical household novel may overfit that novel's failure distribution. The check is now partially executed by the Camel Xiangzi sampling of Section~4: the copied-coordinate failure replicates across corpora (prompt-induced rather than rule-relevant), while the two baseline violations under the primary corpus's criteria and the $26.9\%$ adaptation attrition quantify what genre shift costs the strict-evidence standard; the full-corpus measurement remains the complete check. Parser noise: a wrong temporal or spatial identifier yields a wrong slice even when retrieval and generation are correct; the error-attribution protocol separates parser, retrieval, and generation failures, so a weak parsing result cannot be misread as a retrieval failure, and the lightweight parsing design exists precisely to keep this term small. Cross-genre transfer limits: the ontology of a classical household novel may not cover the relation types of a modern novel, so the generalization experiment reports the adaptation cost rather than assuming zero; the interpretation standard is that transfer holds if accuracy after adaptation is comparable at a bounded rule-engineering cost, and fails if each new genre demands a fresh ontology effort comparable to the original. Cross-language transfer is explicitly out of scope, since it would require re-authoring the ontology and re-engineering extraction prompts, and is left as future work.

\subsection{Case Study: Network Evolution around the Grand View Garden Investigation}
The analysis section also carries the framework's distant-reading demonstration design \cite{moretti2013distant}.

\begin{figure}[ht]\centering
\includegraphics[width=\textwidth]{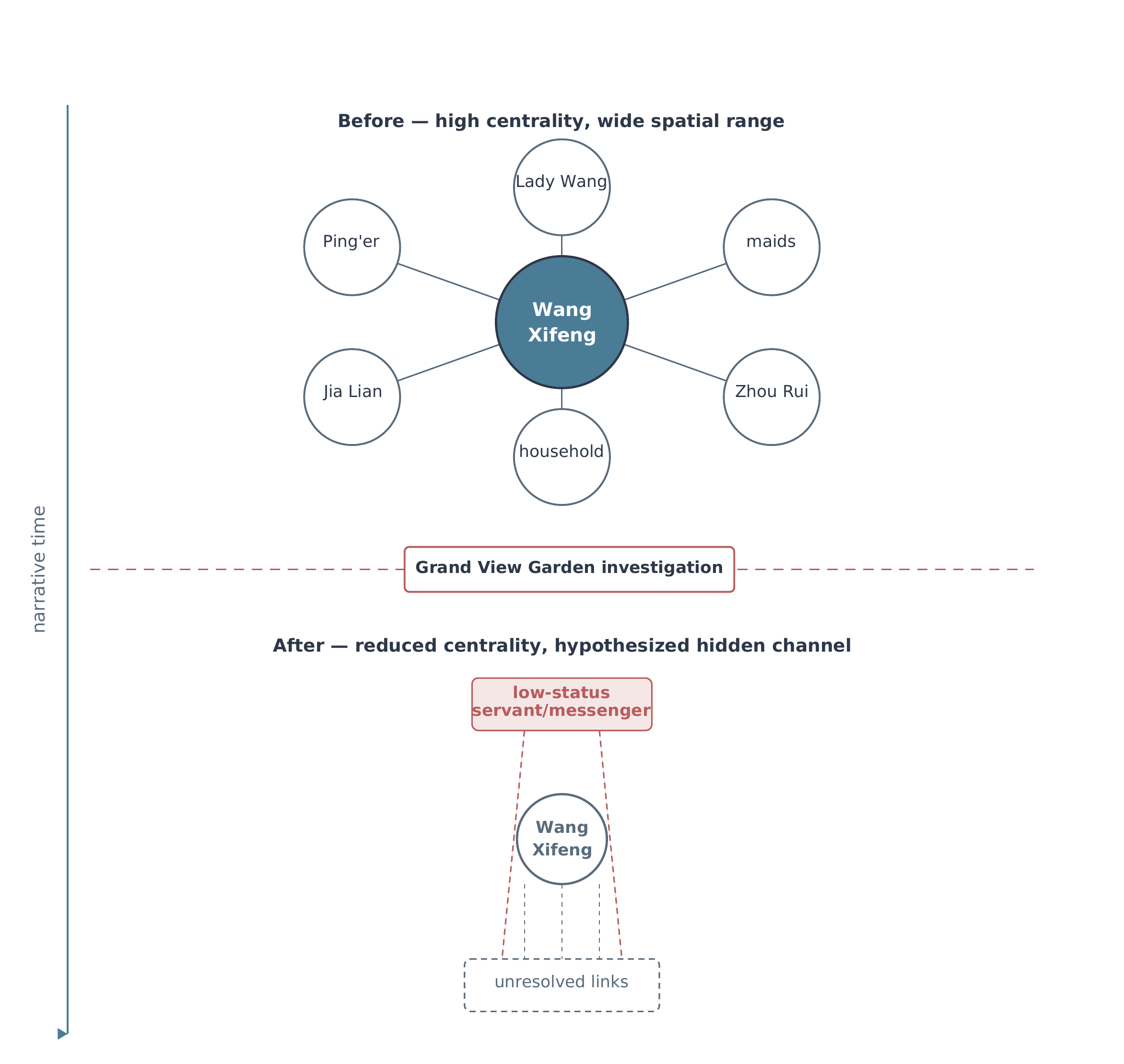}
\caption{Conceptual view of the evolution-analytics case study. The horizontal axis is narrative time; the dashed vertical line marks the Grand View Garden investigation, the dividing event. Before the event, the social network of Wang Xifeng (large central node) shows high centrality and a wide spatial range, drawn as dense connections to household members. After the event, her node is smaller and peripheral. A low-status node (marked ``node?'' in red) is drawn bridging two clusters after the event, representing hypothesis H4b: a hidden information channel maintained by a servant or messenger even as the mistress's own centrality declines. The question mark is deliberate: the figure states an open, falsifiable hypothesis, not a claimed finding; no numeric values appear anywhere in the figure (illustrative only).}\label{fig:centrality_evolution}
\end{figure}

We track the social network of the central character Wang Xifeng across narrative time (Figure~\ref{fig:centrality_evolution}): her degree and betweenness centrality \cite{wasserman1994social,freeman1977betweenness} among the household network, and her spatial activity range across scenes, before and after the investigation of the Grand View Garden, a recognized turning point in the novel's power structure. The analysis protocol is fixed in advance to keep the study from becoming a search for favorable windows: the before and after periods are defined by the event's own narrative-time interval, the centrality measures follow the standard social network analysis definitions \cite{wasserman1994social,freeman1977betweenness}, computed on the versioned slices falling inside each period, and the spatial activity range is defined as the set of scenes in which the character appears with at least one extracted interaction in that period. The analysis has two layers. The first is a validity check: the quantitative trajectory should reproduce the power decline literary scholarship describes, verifying that the temporal graph captures what close reading already knows. The second is a falsifiable hypothesis, H4b: betweenness analysis of low-status nodes may reveal hidden information channels, for instance a servant whose brokerage persists or rises after the investigation even as the mistress's centrality declines. The test is pre-registered: the target set is servant-class nodes appearing in both periods, and the statistic is their maximum brokerage change against a node-permutation null. If the data support H4b, the system surfaces a network mechanism close reading has not formalized; if not, the null result is reported and the case study still demonstrates the evolution-analytics capability. Neither outcome is assumed in advance, and no numbers are reported for the case study in this installment; it is scheduled with the full campaign. The case study also serves a methodological role beyond its literary interest: it exercises the versioned-slice machinery end to end, from narrative-time recovery through entity resolution to evolution queries, on a question class that the benchmark's accuracy metrics do not directly score. A system whose slices are correct enough to reproduce the known power decline is a system whose temporal modeling has passed a real-world check that no synthetic query can replace; a system whose slices are not, fails visibly here even if its QA accuracy is acceptable, because the validity layer and the accuracy layer are measured independently. The relationship to prior character-network work is equally explicit: where earlier studies built static networks from dialogue \cite{elson2010social} or learned static story embeddings \cite{lee2020story}, the versioned slices make the network a time-varying object that can be interrogated meaningfully.

\section{Conclusion}
This paper presented NS-ST-GraphRAG, a framework for knowledge processing over long-form literary narrative built from three coupled components: ontology-guided extraction with a neuro-symbolic constraint layer that rejects and repairs logically impossible relations, a spatio-temporal graph model with dual temporal coordinates and spatial scene attributes, and dynamic sub-graph retrieval designed to answer time- and space-constrained questions from the correct narrative slice with traceable provenance. Together these components give the system an explicit stance on what it knows and when it knows it. The framework is paired with Red-Chamber-QA, to our knowledge the first open multi-hop QA benchmark for classical Chinese literature: its 104-question preliminary set (16 multi-hop), constructed under the mechanical rule set and expert verification, was released with a $0\%$ shortcut rate. The first measured installment characterized three answer-generation baselines and a mechanical oracle ceiling under the dual-track protocol, and implemented the current configuration across all 120 chapters, whose frozen development-set configuration reaches DPC-mech $0.510$ against the window baseline's $0.413$ with citation faithfulness $0.913$ on the shared context track and per-category judge rates $0.861$/$1.000$/$0.533$ (development-set diagnostics): parameter memory alone reached 44.1\% semantic accuracy on judge-scored answers ($0.250$ counting no-verdicts as incorrect) with no source text; the two text-augmented baselines showed overlapping accuracy intervals (DPC-judge 0.915 vs.\ 0.868) while an exploratory pre-freeze zero-shot run on 46 questions scored citation faithfulness 0.109 against 0.510 under the frozen demonstrations; and long-context semantic correctness (0.915) exceeded strict mechanical matching (0.231) by 68.4 percentage points, quantifying the gap the dual-track protocol exposes. The oracle ceiling scores 1.0 on both tracks over its 69 valid verdicts. The held-out campaign then ran once under the frozen configuration over the 120-question test split: the system reproduces answers mechanically on 0.733 against the frozen window baseline's 0.675 and a closed-book floor of 0.083 (McNemar exact p=0.092, not significant), with judge rates 0.866/0.850/0.274 and a difficulty stratification in which the nine hard and four multi-hop questions cut mechanical reproduction to 0.444 and 0.250; H2's constrained-category condition is unmet and disclosed as such. The comparison against the four pre-registered graph-augmented baselines, the extraction-quality, latency, and construction-cost metrics, and the full-corpus generalization measurement remain scheduled. A first cross-genre sampling is already measured: on three chapters of Camel Xiangzi the pipeline transfers with the same schema prompt, the copied-coordinate failure replicates and is repaired deterministically, and after a bounded adaptation (six fixes, seven deletions, one predicate addition, expert-estimated 25--35 minutes) all 19 retained relations pass, a strict hallucination rate of 0\%. The pre-registered case-study design illustrates the framework's potential for distant reading; cross-language transfer remains future work. This installment establishes the evaluation discipline: every number traceable, every answer dual-track scored or counted incorrect under a disclosed bound, every citation mechanically checked for verbatimness.

Declaration of generative AI in scientific writing: the author used generative AI tools to improve the readability and language of the manuscript, reviewed and edited all content, and takes full responsibility for the publication.

\FloatBarrier
\bibliographystyle{unsrtnat}
\bibliography{refs}

@article{shi2023relagraph,
  author={Bin Shi and Hao Wang and Yueyan Li and Sanhong Deng},
  title={RelaGraph: Improving embedding on small-scale sparse knowledge graphs by neighborhood relations},
  journal={Information Processing \& Management},
  year={2023},
  volume={60},
  number={5},
  pages={103447},
  doi={10.1016/j.ipm.2023.103447},
}

@article{yuan2026knowledge,
  author={Wei Yuan and Sha Liu},
  title={Knowledge graph construction for Dream of the Red Chamber through human AI collaboration using large language models},
  journal={Discover Applied Sciences},
  year={2026},
  doi={10.1007/s42452-026-09372-9},
}

@inproceedings{yang2018hotpotqa,
  author={Zhilin Yang and Peng Qi and Saizheng Zhang and Yoshua Bengio and William Cohen and Ruslan Salakhutdinov and Christopher D. Manning},
  title={HotpotQA: A Dataset for Diverse, Explainable Multi-hop Question Answering},
  booktitle={Proceedings of the 2018 Conference on Empirical Methods in Natural Language Processing},
  year={2018},
  pages={2369--2380},
  doi={10.18653/v1/d18-1259},
}

@article{liu2022tlogic,
  author={Yushan Liu and Yunpu Ma and Marcel Hildebrandt and Mitchell Joblin and Volker Tresp},
  title={TLogic: Temporal Logical Rules for Explainable Link Forecasting on Temporal Knowledge Graphs},
  journal={Proceedings of the AAAI Conference on Artificial Intelligence},
  year={2022},
  volume={36},
  number={4},
  pages={4120--4127},
  doi={10.1609/aaai.v36i4.20330},
}

@article{liang2024a,
  author={Ke Liang and Lingyuan Meng and Meng Liu and Yue Liu and Wenxuan Tu and Siwei Wang and Sihang Zhou and Xinwang Liu and Fuchun Sun and Kunlun He},
  title={A Survey of Knowledge Graph Reasoning on Graph Types: Static, Dynamic, and Multi-Modal},
  journal={IEEE Transactions on Pattern Analysis and Machine Intelligence},
  year={2024},
  volume={46},
  number={12},
  pages={9456--9478},
  doi={10.1109/tpami.2024.3417451},
  eprint={2212.05767},
  archivePrefix={arXiv},
}

@article{allen1983maintaining,
  author={James F. Allen},
  title={Maintaining knowledge about temporal intervals},
  journal={Communications of the ACM},
  year={1983},
  volume={26},
  number={11},
  pages={832--843},
  doi={10.1145/182.358434},
}

@article{lee2020story,
  author={O-Joun Lee and Jason J. Jung},
  title={Story embedding: Learning distributed representations of stories based on character networks},
  journal={Artificial Intelligence},
  year={2020},
  volume={281},
  pages={103235},
  doi={10.1016/j.artint.2020.103235},
}

@inproceedings{chen2019understanding,
  author={Jifan Chen and Greg Durrett},
  title={Understanding Dataset Design Choices for Multi-hop Reasoning},
  booktitle={Proceedings of the 2019 Conference of the North American Chapter of the Association for Computational Linguistics: Human Language Technologies, Volume 1 (Long and Short Papers)},
  year={2019},
  pages={4026--4032},
  doi={10.18653/v1/n19-1405},
}

@inproceedings{lewis2020rag,
  author={Patrick Lewis and Ethan Perez and Aleksandra Piktus and Fabio Petroni and Vladimir Karpukhin and Naman Goyal and Heinrich K{\"u}ttler and Mike Lewis and Wen-tau Yih and Tim Rockt{\"a}schel and Sebastian Riedel and Douwe Kiela},
  title={Retrieval-Augmented Generation for Knowledge-Intensive NLP Tasks},
  booktitle={Advances in Neural Information Processing Systems},
  year={2020},
  volume={33},
  pages={9459--9474},
  eprint={2005.11401},
  archivePrefix={arXiv},
}

@article{gao2024survey,
  author={Yunfan Gao and Yun Xiong and Xinyu Gao and Kangxiang Jia and Jinliu Pan and Yuxi Bi and Yi Dai and Jiawei Sun and Meng Wang and Haofen Wang},
  title={Retrieval-Augmented Generation for Large Language Models: A Survey},
  journal={arXiv preprint arXiv:2312.10997},
  year={2024},
  eprint={2312.10997},
  archivePrefix={arXiv},
}

@article{edge2024graphrag,
  author={Darren Edge and Ha Trinh and Newman Cheng and Joshua Bradley and Alex Chao and Apurva Mody and Steven Truitt and Jonathan Larson},
  title={From Local to Global: A Graph RAG Approach to Query-Focused Summarization},
  journal={arXiv preprint arXiv:2404.16130},
  year={2024},
  eprint={2404.16130},
  archivePrefix={arXiv},
}

@article{guo2024lightrag,
  author={Zirui Guo and Lianghao Xia and Yanhua Yu and Tu Ao and Chao Huang},
  title={LightRAG: Simple and Fast Retrieval-Augmented Generation},
  journal={arXiv preprint arXiv:2410.05779},
  year={2024},
  eprint={2410.05779},
  archivePrefix={arXiv},
}

@inproceedings{gutierrez2024hipporag,
  author={Bernal Jim{\'e}nez Guti{\'e}rrez and Yiheng Shu and Yu Gu and Michihiro Yasunaga and Yu Su},
  title={HippoRAG: Neurobiologically Inspired Long-Term Memory for Large Language Models},
  booktitle={Advances in Neural Information Processing Systems},
  year={2024},
  volume={37},
  eprint={2405.14831},
  archivePrefix={arXiv},
}

@inproceedings{sarthi2024raptor,
  author={Parth Sarthi and Salman Abdullah and Aditi Tuli and Shubh Khanna and Anna Goldie and Christopher D. Manning},
  title={RAPTOR: Recursive Abstractive Processing for Tree-Organized Retrieval},
  booktitle={International Conference on Learning Representations},
  year={2024},
  eprint={2401.18059},
  archivePrefix={arXiv},
}

@inproceedings{asai2023selfrag,
  author={Akari Asai and Zeqiu Wu and Yizhong Wang and Avirup Sil and Hannaneh Hajishirzi},
  title={Self-RAG: Learning to Retrieve, Generate, and Critique through Self-Reflection},
  booktitle={International Conference on Learning Representations},
  year={2024},
  eprint={2310.11511},
  archivePrefix={arXiv},
}

@article{huang2024surveyhalluc,
  author={Lei Huang and Weijiang Yu and Weitao Ma and Weihong Zhong and Zhangyin Feng and Haotian Wang and Qianglong Chen and Weihua Peng and Xiaocheng Feng and Bing Qin and Ting Liu},
  title={A Survey on Hallucination in Large Language Models: Principles, Taxonomy, and Open Questions},
  journal={ACM Transactions on Information Systems},
  year={2025},
  volume={43},
  number={2},
  pages={42},
  eprint={2311.05232},
  archivePrefix={arXiv},
}

@inproceedings{min2023factscore,
  author={Sewon Min and Kalpesh Krishna and Xinxi Lyu and Mike Lewis and Wen-tau Yih and Pang Wei Koh and Mohit Iyyer and Luke Zettlemoyer and Hannaneh Hajishirzi},
  title={FActScore: Fine-grained Atomic Evaluation of Factual Precision in Long Form Text Generation},
  booktitle={Proceedings of the 2023 Conference on Empirical Methods in Natural Language Processing},
  year={2023},
  pages={12076--12100},
  eprint={2305.14251},
  archivePrefix={arXiv},
}

@inproceedings{pan2023logiclm,
  author={Liangming Pan and Alon Albalak and Xinyi Wang and William Yang Wang},
  title={Logic-LM: Empowering Large Language Models with Symbolic Solvers for Faithful Logical Reasoning},
  booktitle={Findings of the Association for Computational Linguistics: EMNLP 2023},
  year={2023},
  pages={3806--3824},
  eprint={2305.12295},
  archivePrefix={arXiv},
}

@inproceedings{ho2020constructing,
  author={Xanh Ho and Anh-Khoa Duong Nguyen and Saku Sugawara and Akiko Aizawa},
  title={Constructing A Multi-hop QA Dataset for Comprehensive Evaluation of Reasoning Steps},
  booktitle={Proceedings of the 28th International Conference on Computational Linguistics},
  year={2020},
  pages={6609--6625},
  eprint={2011.01060},
  archivePrefix={arXiv},
}

@article{kocisky2018narrativeqa,
  author={Tomas Kocisky and Jonathan Schwarz and Phil Blunsom and Chris Dyer and Karl Moritz Hermann and Gabor Melis and Edward Grefenstette},
  title={The NarrativeQA Reading Comprehension Challenge},
  journal={Transactions of the Association for Computational Linguistics},
  year={2018},
  volume={6},
  pages={317--328},
  eprint={1712.07040},
  archivePrefix={arXiv},
}

@inproceedings{yang2024crag,
  author={Xiao Yang and Kai Sun and Hao Xin and Yushi Sun and Nikita Bhalla and Xiangsen Chen and Sajal Choudhary and Rongze Daniel Gui and Ziran Will Jiang and Ziyu Jiang and Lingkun Kong and Brian Moran and Jiaqi Wang and Yifan Ethan Xu and An Yan and Chenyu Yang and Eting Yuan and Hanwen Zha and Nan Tang and Lei Chen and Nicolas Scheffer and Yue Liu and Nirav Shah and Rakesh Wanga and Anuj Kumar and Wen-tau Yih and Xin Luna Dong},
  title={CRAG -- Comprehensive RAG Benchmark},
  booktitle={Advances in Neural Information Processing Systems Datasets and Benchmarks Track},
  year={2024},
  eprint={2406.04744},
  archivePrefix={arXiv},
}

@article{peng2024graphragsurvey,
  author={Boci Peng and Yunfan Zhu and Yongchao Liu and Xiaohe Bo and Haizhou Shi and Chuntao Hong and Yan Zhang and Siliang Tang},
  title={Graph Retrieval-Augmented Generation: A Survey},
  journal={arXiv preprint arXiv:2408.08921},
  year={2024},
  eprint={2408.08921},
  archivePrefix={arXiv},
}

@inproceedings{etzioni2011openie,
  author={Oren Etzioni and Anthony Fader and Janara Christensen and Stephen Soderland and Mausam},
  title={Open Information Extraction: The Second Generation},
  booktitle={Proceedings of the Twenty-Second International Joint Conference on Artificial Intelligence},
  year={2011},
  pages={3--10},
  doi={10.5591/978-1-57735-516-8/IJCAI11-012},
}

@inproceedings{huguetcabot2021rebel,
  author={Pere-Llu{\'i}s Huguet Cabot and Roberto Navigli},
  title={REBEL: Relation Extraction By End-to-end Language generation},
  booktitle={Findings of the Association for Computational Linguistics: EMNLP 2021},
  year={2021},
  pages={2370--2381},
  eprint={2109.03227},
  archivePrefix={arXiv},
}

@article{cai2024tkg,
  author={Li Cai and Xin Mao and Yuhao Zhou and Zhaoguang Long and Changxu Wu and Man Lan},
  title={A Survey on Temporal Knowledge Graph: Representation Learning and Applications},
  journal={arXiv preprint arXiv:2403.04782},
  year={2024},
  eprint={2403.04782},
  archivePrefix={arXiv},
}

@article{plamper2025stkg,
  author={Philipp Plamper and Hanna K{\"o}pcke and Anika Gro{\ss}},
  title={A Survey on Spatio-Temporal Knowledge Graph Models},
  journal={arXiv preprint arXiv:2512.16487},
  year={2025},
  eprint={2512.16487},
  archivePrefix={arXiv},
}

@article{ji2021survey,
  author={Shaoxiong Ji and Shirui Pan and Erik Cambria and Pekka Marttinen and Philip S. Yu},
  title={A Survey on Knowledge Graphs: Representation, Acquisition, and Applications},
  journal={IEEE Transactions on Neural Networks and Learning Systems},
  year={2022},
  volume={33},
  number={2},
  pages={494--514},
  doi={10.1109/TNNLS.2021.3070843},
}

@inproceedings{wagner2025mitigating,
  author={Robin Wagner and Emanuel Kitzelmann and Ingo Boersch},
  title={Mitigating Hallucination by Integrating Knowledge Graphs into LLM Inference -- a Systematic Literature Review},
  booktitle={Proceedings of the 63rd Annual Meeting of the Association for Computational Linguistics (Student Research Workshop)},
  year={2025},
  pages={795--805},
  url={https://aclanthology.org/2025.acl-srw.53/},
}

@inproceedings{elson2010social,
  author={David K. Elson and Nicholas Dames and Kathleen R. McKeown},
  title={Extracting Social Networks from Literary Fiction},
  booktitle={Proceedings of the 48th Annual Meeting of the Association for Computational Linguistics},
  year={2010},
  pages={138--147},
  url={https://aclanthology.org/P10-1015/},
}

@book{moretti2013distant,
  author={Franco Moretti},
  title={Distant Reading},
  publisher={Verso Books},
  address={London},
  year={2013},
  isbn={978-1-78168-084-1},
  url={https://find.library.duke.edu/catalog/DUKE006035194},
}

@article{zhang2026preqin,
  author={Qiang Zhang and Shupeng Guan and Fanghong Xu},
  title={Research on the Construction of Knowledge Graphs for Pre-Qin Canonical Texts Driven by Agent Technologies},
  journal={Journal of Nantong University (Social Sciences Edition)},
  year={2026},
  volume={42},
  number={1},
  pages={142--154},
  url={https://www.cqvip.com/doc/journal/7202877532},
}

@article{wei2025daguanyuan,
  author={Shilong Wei and Xiaojun Sheng and Minmin Li},
  title={Graph-semantics-driven 3D spatial reconstruction of historical cultural heritage: a case study of the Grand View Garden in Dream of the Red Chamber},
  journal={Journal of Natural Resources},
  year={2025},
  volume={40},
  number={12},
}

@inproceedings{huang2023tgb,
  author={Shenyang Huang and Farimah Poursafaei and Jacob Danovitch and Matthias Fey and Weihua Hu and Emanuele Rossi and Jure Leskovec and Michael Bronstein and Guillaume Rabusseau and Reihaneh Rabbany},
  title={Temporal Graph Benchmark for Machine Learning on Temporal Graphs},
  booktitle={Advances in Neural Information Processing Systems Datasets and Benchmarks Track},
  year={2023},
  eprint={2307.01026},
  archivePrefix={arXiv},
}

@inproceedings{trivedi2017knowevolve,
  author={Rakshit Trivedi and Hanjun Dai and Yichen Wang and Le Song},
  title={Know-Evolve: Deep Temporal Reasoning for Dynamic Knowledge Graphs},
  booktitle={Proceedings of the 34th International Conference on Machine Learning},
  year={2017},
  pages={3462--3471},
  eprint={1705.05742},
  archivePrefix={arXiv},
}

@inproceedings{li2021regen,
  author={Zixuan Li and Xiaolong Jin and Wei Li and Saiping Guan and Jiafeng Guo and Huawei Shen and Yuanzhuo Wang and Xueqi Cheng},
  title={Temporal Knowledge Graph Reasoning Based on Evolutional Representation Learning},
  booktitle={Proceedings of the 44th International ACM SIGIR Conference on Research and Development in Information Retrieval},
  year={2021},
  pages={408--417},
  eprint={2104.10353},
  archivePrefix={arXiv},
}

@article{sadeghi2021chronor,
  author={Ali Sadeghian and Mohammadreza Armandpour and Anthony Colas and Daisy Zhe Wang},
  title={ChronoR: Rotation Based Temporal Knowledge Graph Embedding},
  journal={Proceedings of the AAAI Conference on Artificial Intelligence},
  year={2021},
  volume={35},
  number={7},
  pages={6471--6479},
  doi={10.1609/aaai.v35i7.16802},
}

@inproceedings{vala2015mrbennet,
  author={Hannah Vala and David Jurgens and Andrew Piper and Derek Ruths},
  title={Mr. Bennet, his coachman, and the Archbishop walk into a bar but only one of them gets recognized: On The Difficulty of Detecting Characters in Literary Texts},
  booktitle={Proceedings of the 2015 Conference on Empirical Methods in Natural Language Processing},
  year={2015},
  pages={769--774},
  url={https://aclanthology.org/D15-1055/},
}

@article{freeman1977betweenness,
  author={Linton C. Freeman},
  title={A Set of Measures of Centrality Based on Betweenness},
  journal={Sociometry},
  year={1977},
  volume={40},
  number={1},
  pages={35--41},
  doi={10.2307/3033543},
}

@book{wasserman1994social,
  author={Stanley Wasserman and Katherine Faust},
  title={Social Network Analysis: Methods and Applications},
  publisher={Cambridge University Press},
  address={Cambridge},
  year={1994},
  isbn={0-521-38707-8},
  doi={10.1017/CBO9780511815478},
}

@inproceedings{chambers2008narrative,
  author={Nathanael Chambers and Dan Jurafsky},
  title={Unsupervised Learning of Narrative Event Chains},
  booktitle={Proceedings of ACL-08: HLT},
  year={2008},
  pages={789--797},
  url={https://aclanthology.org/P08-1090/},
}

@techreport{page1999pagerank,
  author={Lawrence Page and Sergey Brin and Rajeev Motwani and Terry Winograd},
  title={The PageRank Citation Ranking: Bringing Order to the Web},
  howpublished={Technical Report, Stanford InfoLab},
  year={1999},
  url={http://ilpubs.stanford.edu:8090/422/},
}

\end{document}